\documentclass{article}
\usepackage[preprint]{neurips_2025}
\usepackage[T1]{fontenc}
\usepackage[utf8]{inputenc}

\usepackage{amsmath, amssymb}
\usepackage{graphicx}
\usepackage{float}
\usepackage{caption}
\usepackage{listings}
\usepackage[most]{tcolorbox}
\usepackage{booktabs}
\usepackage{mathtools}
\usepackage{url}

\usepackage{natbib}
\title{\fontsize{16}{21}\selectfont Molecular Representations for Large Language Models}

\author{%
  Nicholas T. Runcie \\
  Department of Statistics \\
  University of Oxford \\
  Oxford, UK \\
  \texttt{runcie@stats.ox.ac.uk} \\
  \And
  Fergus Imrie \\
  Department of Statistics \\
  University of Oxford \\
  Oxford, UK \\
  \texttt{imrie@stats.ox.ac.uk} \\
  \And
  Charlotte M. Deane \\
  Department of Statistics \\
  University of Oxford \\
  Oxford, UK \\
  \texttt{deane@stats.ox.ac.uk} \\
}

\begin{document}

\maketitle

\begin{abstract}
Large Language Models (LLMs) are increasingly being used to support scientific discovery.  
In chemistry, tasks such as reaction prediction and structure elucidation require reasoning about the structures of molecules. 
As such, LLM-based systems for chemistry must interact reliably with molecular structures. 
Most previous studies of LLMs in chemistry have used SMILES strings or IUPAC names as molecular representations; however, the suitability of these formats has not been systematically assessed. 
In this work, we introduce MolJSON, a novel molecular representation for LLMs, and systematically compare it with five common chemical formats. 
We evaluated each representation with GPT-5-nano, GPT-5-mini, GPT-5, and Claude Haiku 4.5 using a set of 78,045 questions spanning translation, shortest path, and constrained generation reasoning tasks. 
We observed substantial variation across representations in the ability of LLMs to interpret and generate molecular graphs, with MolJSON consistently outperforming existing formats.
On translation tasks, GPT-5 achieved 71.0\% accuracy when converting IUPAC names to MolJSON, compared with 43.7\% when converting the same inputs to SMILES.
For constrained generation, GPT-5 reached 95.3\% accuracy generating MolJSON, compared with 76.3\% for IUPAC and 64.0\% for SMILES. 
As an input format for shortest-path reasoning, GPT-5 successfully answered 98.5\% of questions with MolJSON, compared with 92.2\% for SMILES and 82.7\% for IUPAC, whilst also using fewer reasoning tokens. 
We observed systematic errors associated with atom count and ring complexity for SMILES strings and IUPAC names, whereas MolJSON was more robust to these failure modes.
Our results show that the choice of molecular representation has a material impact on LLM performance, and that explicit molecular graph schemas, such as MolJSON, are a promising direction for LLM-based systems in chemistry. 
\end{abstract}

\section{Introduction}
Large Language Models (LLMs) are being adopted for a growing range of scientific tasks \citep{alampara_general_2025,mitchener_kosmos_2025,m_bran_augmenting_2024,gottweis_towards_2025,boiko_autonomous_2023}.
To support discovery in chemistry, autonomous scientific systems will need to reason about molecular structures in open-ended settings where standard procedural solutions do not exist, such as molecular design \citep{janet_artificial_2023}, mechanistic analysis \citep{ben-tal_mechanistic_2022,bran_chemical_2025}, and structure elucidation \citep{priessner_enhancing_2026}. 
For LLMs to be used in these applications, molecular structures need to be represented in a format that the models can process. 
Molecules are naturally described as molecular graphs in which atoms are nodes and bonds are edges; 
however, since current LLMs are sequence-based models that primarily operate on text, the molecular graphs must be serialised into a text-based format. 
It is likely that the choice of text-based representation will significantly affect the performance of LLMs on chemistry tasks. 

Previous studies of LLMs in chemistry have used existing text-based molecular representations, predominantly SMILES strings \citep{ChemIQ,bran_chemical_2025,guo_what_2023,Ether0} and IUPAC names \citep{sharlin_nmr_challenge_2025}.
However, the choice of molecular representation provided to LLMs has been shown to materially affect their performance on chemistry-based tasks \citep{yan_inconsistency_2025,ChemIQ}.
A likely reason for the varied performance is that text-based molecular formats impose syntactic and semantic constraints on how structures are expressed.
While these formats are widely used in cheminformatics, they were designed for traditional computational workflows and human interpretation rather than current LLM-based systems. 
As such, this motivates the exploration of molecular representations for LLM-based reasoning.

To address this need, we introduce MolJSON, a structured JSON schema for molecular graphs, as an alternative molecular representation specifically designed for LLMs.
MolJSON encodes molecules as JSON objects \citep{bray2014json} using explicit lists of atoms and bonds, where atoms are specified by unique identifiers and elements, and bonds by pairs of atom identifiers and a bond order. 
Since molecular graphs are defined by the set of atoms and bonds, this eliminates the need to linearise graphs into a specific traversal (as required for SMILES strings), or to encode them through rule-based nomenclature (as with IUPAC names). 
Additionally, MolJSON is designed to be compatible with the structured output mode of LLMs, in which responses conform to a predefined JSON schema \citep{We_Need_Structured_Output,geng2025jsonschemabench}. 
This mode, which is now supported by many LLMs, allows responses to be parsed by external software and is commonly used in agentic systems. 

We systematically evaluate the ability of general-purpose LLMs to interpret and generate molecular structures across five common chemical representations and our proposed MolJSON format.
We assessed GPT-5-nano, GPT-5-mini, GPT-5, and Claude Haiku 4.5 using 78,045 algorithmically generated questions spanning translation, shortest-path reasoning, and constrained molecular generation. 
These tasks were designed to isolate the ability of LLMs to interact with molecular structures, rather than to assess performance on applied chemistry tasks.
We observed significant variability in performance across all tasks depending on the molecular representation tested, with MolJSON consistently outperforming the standard formats. 
For example, on translation questions, GPT-5 achieved 71.0\% accuracy converting IUPAC to MolJSON, compared with 43.7\% for IUPAC to SMILES, with similar improvements for the shortest path and constrained generation tasks. 
Despite the ubiquity of SMILES strings and IUPAC names in chemical databases and their presence in LLM training data, MolJSON outperformed both formats without any explicit training.
Together, these results show that the choice of molecular representation materially affects how reliably LLMs can interpret and generate molecular graphs, and suggest that explicit graph encodings, such as MolJSON, are a promising direction for LLM-based systems in chemistry.

\section{Methods}

\begin{figure}[p]
  \centering

  \noindent
  \begin{tcolorbox}[
    enhanced,
    colback=white,
    colframe=white,
    boxrule=0pt,
    sidebyside,
    sidebyside align=top,
    segmentation style={draw=none},
    righthand width=0.4\linewidth,
    sidebyside gap=0.02\linewidth,
    left=0pt,
    right=0pt,
    top=0pt,
    bottom=0pt,
    boxsep=0pt
  ]
    \begin{tcolorbox}[
      colback=white,
      colframe=white,
      boxrule=0pt,
      equal height group=ethanoic-panels,
      height fixed for=all,
      left=0pt,
      right=0pt,
      top=0pt,
      bottom=0pt,
      boxsep=0pt
    ]
      \begin{tcolorbox}[
        colback=white,
        colframe=black,
        boxrule=0.5pt,
        arc=0pt,
        left=4pt,
        right=4pt,
        top=4pt,
        bottom=4pt,
        fontupper=\ttfamily,
        title={IUPAC}
      ]
      ethanoic acid
      \end{tcolorbox}

      \begin{tcolorbox}[
        colback=white,
        colframe=black,
        boxrule=0.5pt,
        arc=0pt,
        left=4pt,
        right=4pt,
        top=4pt,
        bottom=4pt,
        fontupper=\ttfamily,
        title={InChI}
      ]
      InChI=1S/C2H4O2/c1-2(3)4/h1H3,(H,3,4)
      \end{tcolorbox}

      \begin{tcolorbox}[
        colback=white,
        colframe=black,
        boxrule=0.5pt,
        arc=0pt,
        left=4pt,
        right=4pt,
        top=4pt,
        bottom=4pt,
        fontupper=\ttfamily,
        title={SMILES}
      ]
      CC(=O)O
      \end{tcolorbox}

      \begin{tcolorbox}[
        colback=white,
        colframe=black,
        boxrule=0.5pt,
        arc=0pt,
        left=4pt,
        right=4pt,
        top=4pt,
        bottom=4pt,
        fontupper=\ttfamily,
        title={SELFIES}
      ]
      [O][C][=Branch1][C][=O][=C]
      \end{tcolorbox}
    \end{tcolorbox}

    \tcblower

    \begin{tcolorbox}[
      colback=white,
      colframe=black,
      boxrule=0.5pt,
      arc=0pt,
      left=4pt,
      right=4pt,
      top=4pt,
      bottom=4pt,
      fontupper=\ttfamily,
      title={Skeletal Diagram},
      equal height group=ethanoic-panels,
      height fixed for=all,
      valign=center
    ]
    \centering
    \includegraphics[width=0.9\linewidth]{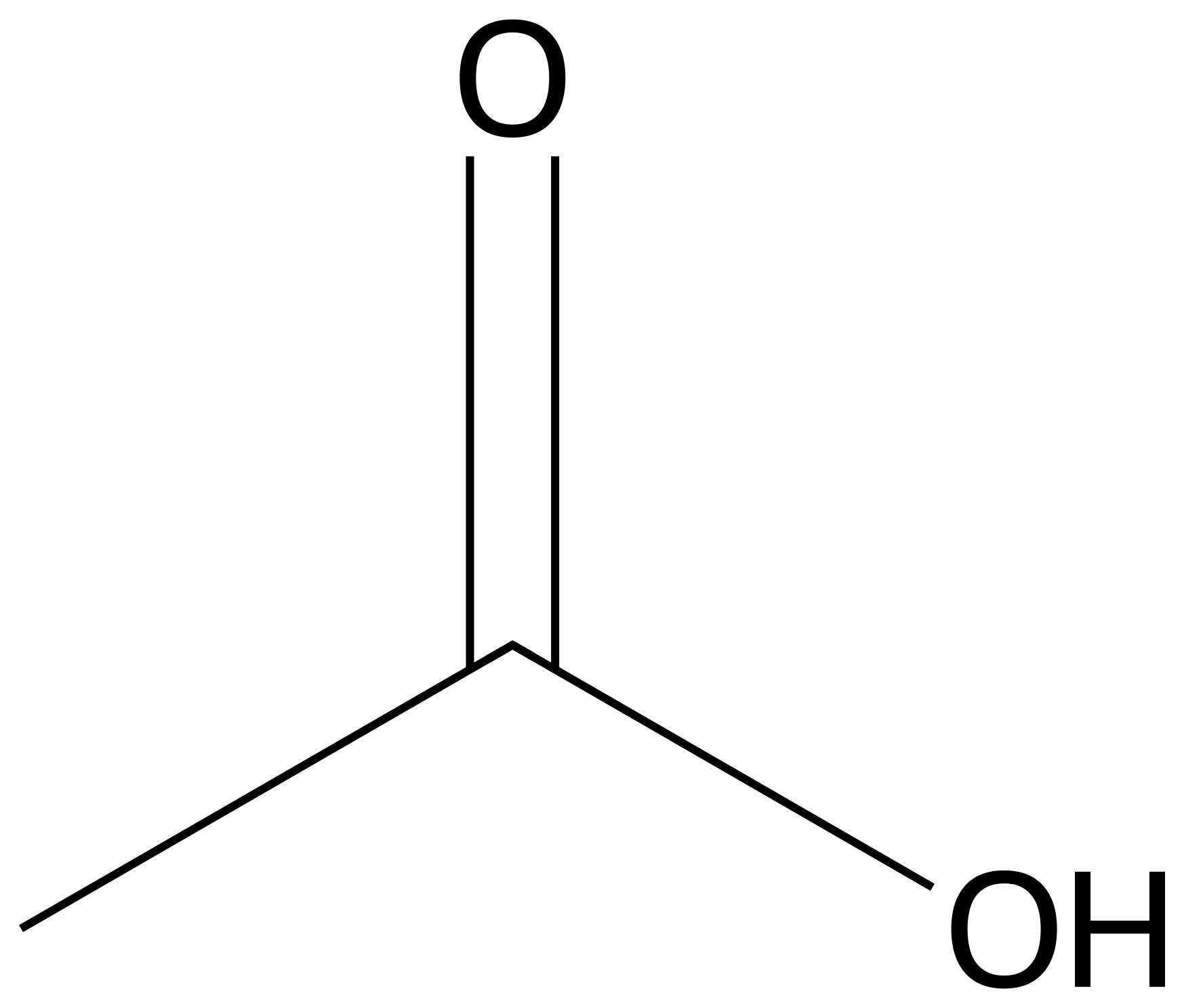}
    \end{tcolorbox}
  \end{tcolorbox}

  \begin{tcolorbox}[
    colback=white,
    colframe=black,
    boxrule=0.5pt,
    arc=0pt,
    left=4pt,
    right=4pt,
    top=4pt,
    bottom=4pt,
    fontupper=\ttfamily,
    title={MOL V2000}
  ]
\begin{lstlisting}[
  basicstyle=\ttfamily\normalsize,
  frame=none,
  breaklines=false,
  showstringspaces=false
]
     RDKit          2D

  4  3  0  0  0  0  0  0  0  0999 V2000
    0.0000   0.0000   0.0000 C  0 0 0 0 0 0 0 0 0 0 0 0
    1.2990   0.7500   0.0000 C  0 0 0 0 0 0 0 0 0 0 0 0
    1.2990   2.2500   0.0000 O  0 0 0 0 0 0 0 0 0 0 0 0
    2.5981  -0.0000   0.0000 O  0 0 0 0 0 0 0 0 0 0 0 0
  1  2  1  0
  2  3  2  0
  2  4  1  0
M  END
\end{lstlisting}
  \end{tcolorbox}

  \begin{tcolorbox}[
    colback=white,
    colframe=black,
    boxrule=0.5pt,
    arc=0pt,
    left=4pt,
    right=4pt,
    top=4pt,
    bottom=4pt,
    fontupper=\ttfamily,
    title={MolJSON (Ours)}
  ]
\begin{lstlisting}[
  basicstyle=\ttfamily\normalsize,
  frame=none,
  breaklines=true,
  breakatwhitespace=false,
  showstringspaces=false
]
{
  "atoms": [
    {"id": "C1", "element": "C"},
    {"id": "C2", "element": "C"},
    {"id": "O1", "element": "O"},
    {"id": "O2", "element": "O"}
  ],
  "bonds": [
    {"source": "C1", "target": "C2", "order": 1},
    {"source": "C2", "target": "O1", "order": 2},
    {"source": "C2", "target": "O2", "order": 1}
  ],
  "charges": null,
  "aromatic_n_h": null
}
\end{lstlisting}
  \end{tcolorbox}

  \caption{The molecular representations considered in this work, illustrated using acetic acid as an example. The skeletal diagram is provided for clarity and was not used as a molecular representation when assessing the LLMs.}
  \label{fig:ethanoic-acid-reprs-panels-ethanoic}
\end{figure}

\subsection{Existing Molecular Representations}
We consider five existing molecular representations in this work; an example of each is shown in Figure \ref{fig:ethanoic-acid-reprs-panels-ethanoic}.
SMILES \citep{weininger_smiles_1988} encodes molecules as text strings by traversing the molecular graph, writing atom and bond symbols in a sequence with parentheses denoting branches and ring numbers indicating where the sequence reconnects with an earlier atom. 
SMILES is ubiquitous in chemical databases \citep{kim_pubchem_2025} and has been widely adopted in task-specific chemical language models, such as for \textit{de novo} molecular design \citep{loeffler_reinvent_2024} and reaction prediction \citep{schwaller_molecular_2019}.
SELFIES \citep{Krenn_2020} is also a traversal-based encoding, but is defined using a decoding scheme that enforces chemical constraints, making any sequence of tokens valid. 
This format was introduced to support robust molecular generation, but has often underperformed SMILES in chemical language models \citep{skinnider_invalid_2024} and has not been used in many other contexts.
IUPAC naming \citep{favre_nomenclature_2013} uses nomenclature rules to generate systematic names of molecules. 
These are built by defining a parent structure and then using locants, prefixes, and suffixes to specify the identity and position of structural features in the molecule. 
IUPAC names are widely used in the scientific literature, patents, and databases as a human-readable molecular representation \citep{papadatos_surechembl_2016}.
InChI \citep{heller_inchi_2015} is designed to represent each molecule as a canonical string. 
These are constructed using a series of string ``layers'' which describe different aspects of the molecule. 
InChI is primarily used to uniquely identify molecules in databases.
Finally, MOL V2000 \citep{mdl_ctfile_formats_2003} is a connection-table file format that stores molecules as explicit blocks of text, with atoms and bonds listed separately and bonds specified by pairs of indices that reference atoms in the atom block.
This format is commonly used for exchanging molecular structures together with their atom coordinates.

\subsection{MolJSON Schema Definition}
\label{sec:MolJSON-schema-def}
MolJSON is our proposed structured JSON schema \citep{pezoa2016foundations} for representing molecular graphs, specifically designed as both an input and output format for LLMs.
Each molecule is represented by two components: an \texttt{"atoms"} array, where each entry specifies a unique identifier and element symbol, and a \texttt{"bonds"} array, where each entry specifies a pair of atom identifiers and the corresponding bond order.
We additionally include two optional sparse fields to ensure correct valence assignment: \texttt{"charges"}, which lists non-zero formal charges, and \texttt{"aromatic\_n\_h"}, which records explicit hydrogens on aromatic nitrogen atoms. 
All remaining hydrogens are treated as implicit and inferred from the graph using standard valence rules.
Molecular identity is defined directly by the sets of atoms and bonds, independent of array ordering, as opposed to traversal-dependent encodings such as SMILES strings or rule-based nomenclature such as IUPAC names.
An example of the MolJSON format for acetic acid is shown in Figure \ref{fig:ethanoic-acid-reprs-panels-ethanoic}, along with the other molecular representations considered in this work. 
The full MolJSON schema specification and accompanying scripts are available in our GitHub repository at \url{https://github.com/oxpig/MolJSON}.

\subsection{Benchmark Design and Task Construction}
We constructed a set of tasks designed to isolate the impact of molecular representation on LLM performance. 
We selected three tasks: molecular translation, shortest-path reasoning, and constrained generation. 
The translation task assesses whether an LLM can read a molecular structure in one format and rewrite the same structure in another; the shortest-path task evaluates whether the LLM can navigate substructures within a molecule; and the constrained generation task assesses whether the LLM can generate a molecule satisfying a set of constraints, serving as a proxy for tasks like structure elucidation. 
These questions enable us to systematically assess individual strengths and weaknesses of each format without performance being obscured by effects relating to specialised chemistry knowledge. 

To generate our questions, we used molecules deposited in the PubChem database between the dates 1 October 2025 and 22 December 2025 \citep{kim_pubchem_2025}.
This range postdates the release of the GPT-5 models, reducing the risk of contamination from PubChem-derived training data, although it does not guarantee their exclusion from training datasets derived from other sources. 
Molecules containing stereochemistry, salts, multiple molecules, isotopic information, radicals, and inorganic elements were excluded from the dataset. 

The first task was molecular translation, which jointly evaluated input and output representations.
For these, the LLM was given a molecule in one format and was prompted to convert it to another, with accuracy defined as exact molecular equivalence between the two structures. 
Molecules were selected by stratified sampling across heavy-atom counts (10-30) and ring counts (0-3). 
Two sets of molecules were selected: a ``small'' set ($n=420$) with five neutral molecules per stratum, and a ``large'' set ($n=9{,}201$) with up to 100 neutral and 10 charged molecules per stratum. 
For the small set, questions were generated using SMILES, IUPAC, MOL V2000, SELFIES, InChI, and MolJSON representations, yielding $12{,}600$ prompts. 
Following the results on this set, we chose to focus all subsequent experiments on SMILES, IUPAC, and MolJSON.
We enumerated translation questions using these formats with the large set of molecules, yielding $55{,}206$ prompts.
Outputs were parsed with RDKit \citep{rdkit} and evaluated by comparison of canonical SMILES to the ground-truth molecule. 
For IUPAC outputs, predicted names were converted to SMILES using the OPSIN parsing tool \citep[version 2.9.0;][]{lowe_chemical_2011} before RDKit evaluation.

To assess input representations in isolation, we used a variation of the shortest-path task introduced in ChemIQ \citep{ChemIQ}.
We selected molecules containing exactly two halogen atoms and prompted the LLM to determine the number of bonds on the shortest path between them. 
Molecules containing exactly two halogen atoms were selected from the PubChem dataset, the shortest path between these two atoms was computed, and we sampled up to 200 molecules for each path length from two to 18 bonds ($n=2{,}919$ molecules, Appendix \ref{sec:shortest-path-length-counts} Table \ref{tab:shortest-path-length-counts}). 
Prompts were generated using SMILES, IUPAC, and MolJSON representations, yielding $8{,}757$ questions. 

Finally, to assess the output representations in isolation, we introduced the constrained molecular generation task in which the LLM was prompted to generate a molecule that satisfied a set of constraints. 
Each question required the presence of one fluorine, chlorine, and bromine atom, together with constraints on the shortest path between each pair of halogen atoms, and on ring counts and topologies. 
Generated molecules were evaluated using RDKit to check if they satisfied these constraints. 
We generated $494$ constraint sets and evaluated SMILES, IUPAC, and MolJSON outputs for each, yielding a total of $1{,}482$ prompts.
Further details and an example of this task are provided in Appendix \ref{app:constrained-mol-gen-methods}. 

\subsection{Model Configuration and Prompting}
The majority of our evaluation was performed using OpenAI models, specifically GPT-5-nano, GPT-5-mini, and GPT-5 \citep{OpenAI_gpt5_system_card}, which span a range of model sizes and performance levels.
Prompts were submitted using the OpenAI Responses API with ``low'' reasoning effort, and no tool calling.
We additionally ran a subset of prompts using Claude Haiku 4.5 \citep{anthropic2025haiku45systemcard}, which were submitted to the Anthropic API with a thinking budget of $4{,}096$ tokens and an output limit of $16{,}384$ tokens. 
The LLMs were run in structured output mode, with JSON schemas defining the response format. 
In cases where the expected output was MolJSON, the MolJSON schema in Appendix \ref{app:schema-full} was provided.
In all other cases, a single-key JSON schema was used where the key was the expected output representation, for example \texttt{\{"smiles":"CCO"\}}. 
Our entire evaluation comprised $78{,}045$ unique questions, and we collected $224{,}055$ model responses in total. 

\begin{figure}[t]
  \centering
  \includegraphics[width=0.7\linewidth]{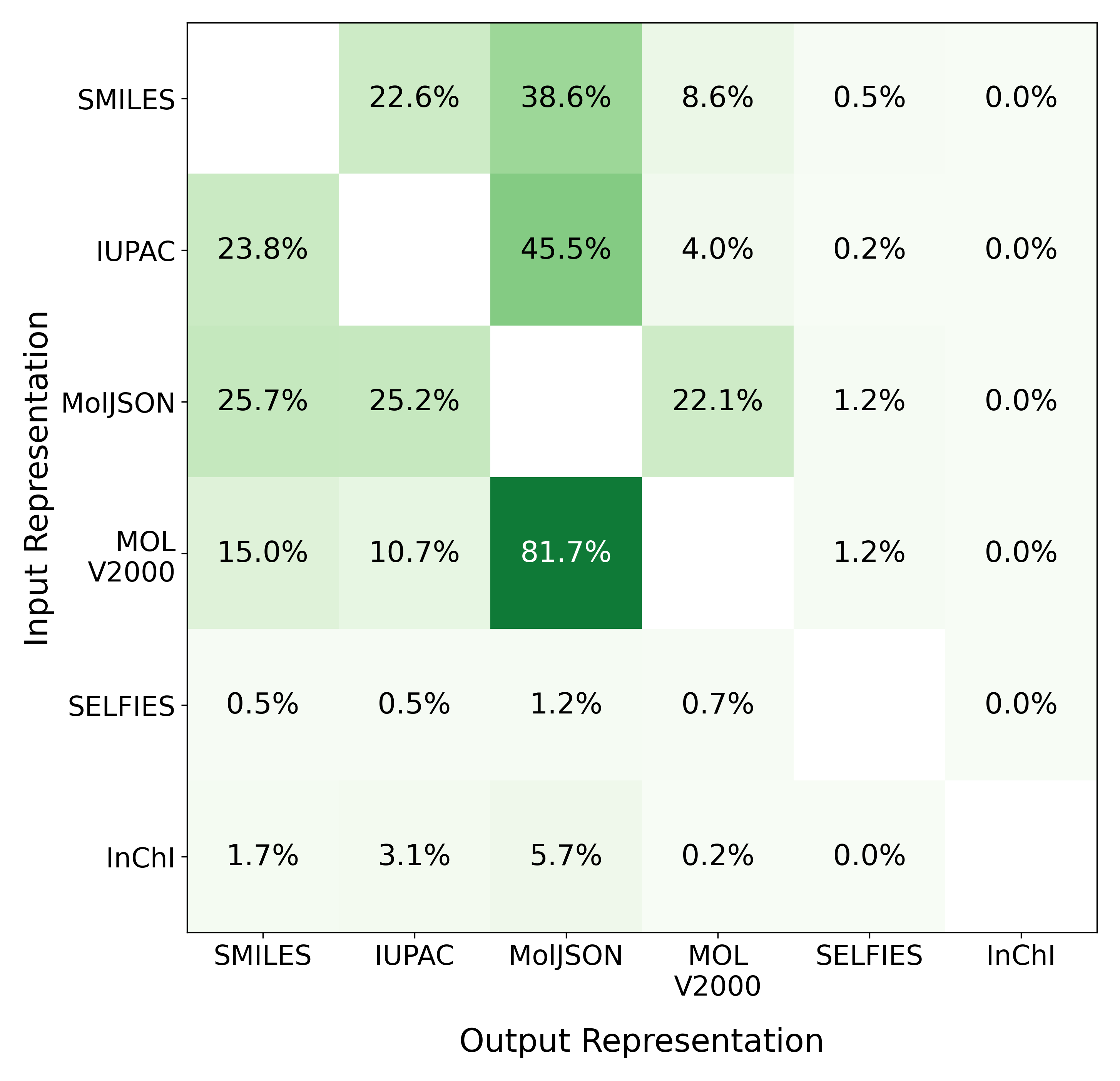}
  \caption{Accuracy of GPT-5-mini on pairwise translation tasks between SMILES, IUPAC, MolJSON, MOL V2000, SELFIES, and InChI. 
  Questions were constructed using $420$ unique molecules sampled from PubChem, yielding a total of $12{,}600$ prompts.}
  \label{fig:gpt5mini-translation-matrix-extended}
\end{figure}

\section{Results and Discussion}

\subsection{Molecular Representation Strongly Influences LLM Performance}
\label{sec:general-translation}

We first assessed the performance of GPT-5-mini in reading and writing molecular representations through pairwise translation tasks with six formats (SMILES, IUPAC, MOL V2000, InChI, SELFIES, and MolJSON). 
Across these, MolJSON, SMILES, and IUPAC were the most consistent formats, with pairwise translation accuracy between them ranging from 22.6\% to 45.5\% (Figure \ref{fig:gpt5mini-translation-matrix-extended}).
MOL V2000 exhibited strong asymmetric performance, with MOL V2000 to MolJSON achieving 81.7\% accuracy, whilst the reverse direction was 22.1\%. 
This is likely due to trivial reformatting of the atom and bond blocks when converting from the MOL V2000 format, rather than being a robust molecular representation.
We suspect the poor performance of SELFIES is due, in part, to this representation being uncommon in public datasets, while for InChI we believe the intrinsic complexity of the format is likely driving the low accuracy. 
A detailed analysis of the failure modes of MOL V2000, SELFIES, and InChI is provided in Appendix \ref{app:mol-v2000}. 
Based on these initial results, we focused the remaining experiments on SMILES, IUPAC, and MolJSON. In the following sections, we assess their performance as input and output representations, before systematically analysing the error modes associated with these three formats in Section \ref{sec:error-analysis}.

\subsection{MolJSON Outperforms SMILES and IUPAC in Molecular Translation Tasks}
\label{sec:translation-smiles-iupac-moljson}

\begin{figure}[t]
  \centering
  \includegraphics[width=\linewidth]{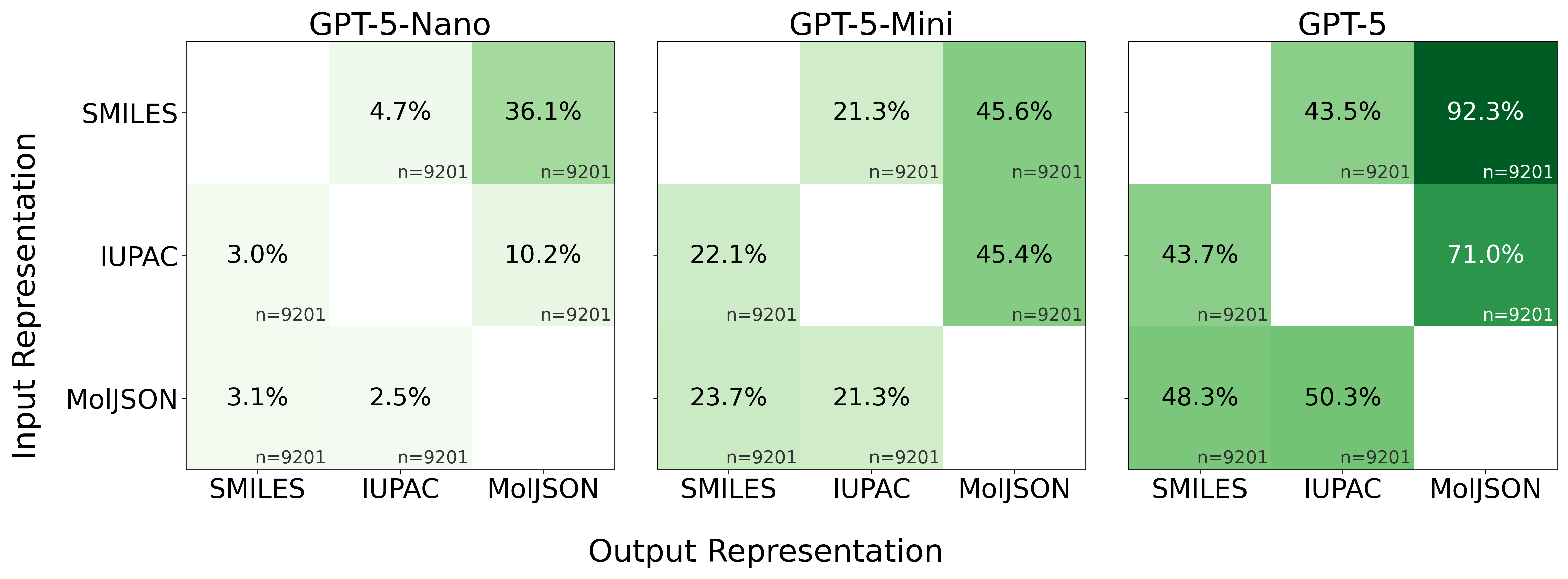}
  \caption{Accuracy of GPT-5-nano, GPT-5-mini, and GPT-5 on pairwise translation between SMILES, IUPAC, and MolJSON representations. Questions were constructed using $9{,}201$ unique molecules sampled from PubChem.}
  \label{fig:translation-matrix-set}
\end{figure}

We compared SMILES, IUPAC, and MolJSON using pairwise translation tasks with 9,201 molecules (55,206 questions total). 
Across GPT-5-nano, GPT-5-mini, and GPT-5, translation accuracy depended most strongly on the output representation, with translations to MolJSON achieving approximately twice the accuracy of the corresponding translations to SMILES or IUPAC (Figure \ref{fig:translation-matrix-set}).
The same trend was also observed for Claude Haiku 4.5 on the smaller set of 420 molecules (2,520 questions, Appendix \ref{sec:haiku-4-5-translation} Figure \ref{fig:haiku-4-5-translation}).

For GPT-5, IUPAC to MolJSON achieved 71.0\% accuracy compared with 43.7\% for IUPAC to SMILES. Likewise, translation from SMILES to MolJSON achieved 92.3\% accuracy compared with 43.5\% for SMILES to IUPAC. 
These results indicate LLMs can often recover the underlying molecule from SMILES and IUPAC inputs, but are substantially less able to express the molecular graph as a SMILES string or IUPAC name, compared with the MolJSON format. 

MolJSON also showed improved accuracy as an input format for GPT-5.
When generating SMILES outputs, accuracy increased from 43.7\% (IUPAC input) to 48.3\% (MolJSON input), and for IUPAC outputs, accuracy improved from 43.5\% (SMILES input) to 50.3\% (MolJSON input).
This modest improvement is consistent with SMILES and IUPAC generation being the limiting factor in translation accuracy, and suggests MolJSON is at least as effective an input representation as SMILES or IUPAC.

\subsection{MolJSON is the Best Input Representation for Shortest-Path Reasoning}

\begin{figure}[t]
  \centering
  \includegraphics[width=0.8\linewidth]{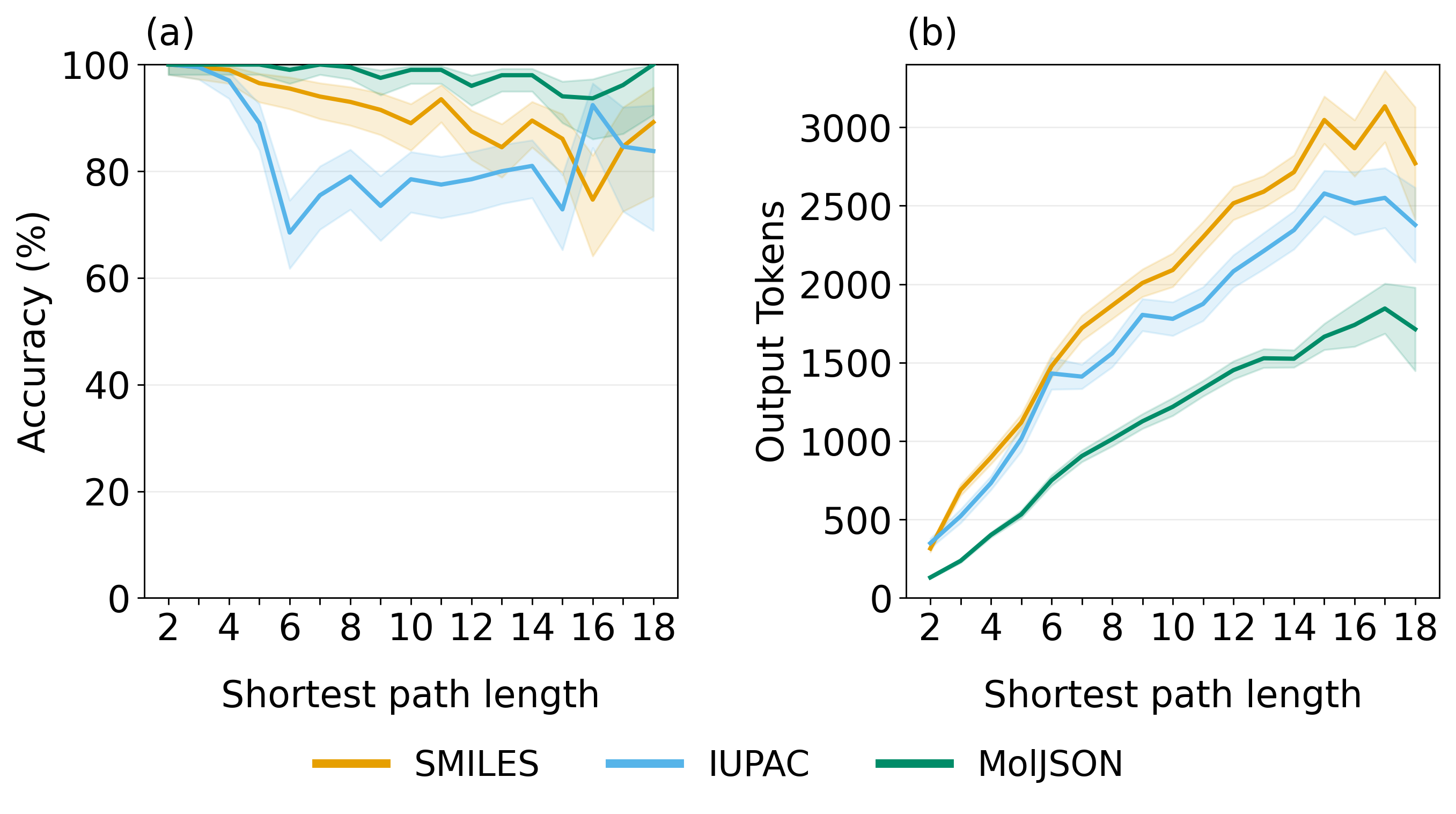}
    \caption{Performance of GPT-5 on the shortest-path task using SMILES, IUPAC, and MolJSON as input representations. (a) Accuracy by path length, with shaded regions indicating 95\% Wilson confidence intervals. (b) Mean number of output tokens by path length, with shaded regions indicating 95\% confidence intervals computed as the mean $\pm 1.96 \times$ the standard error of the mean.}
  \label{fig:gpt5-shortest-path}
\end{figure}

To determine the effect of the choice of input representation, we used shortest-path tasks in which the model was prompted to count the number of bonds along the shortest path between two halogen atoms in a molecule.
We used a set of 2,919 molecules and tested SMILES, IUPAC, and MolJSON formats, yielding 8,757 questions. 
MolJSON was the most reliable input format across all models (Appendix \ref{sec:shortest_path_task_results} Table \ref{tab:shortest_path_by_input}); for GPT-5, MolJSON achieved 98.5\% accuracy, outperforming SMILES, which correctly answered 92.2\% of questions, and IUPAC, which scored 82.7\%, across all path lengths (Figure \ref{fig:gpt5-shortest-path}a). 
The lower performance of IUPAC was largely driven by the presence of fused ring systems (Appendix \ref{sec:shortest_path_task_results} Table \ref{tab:shortest-path-fused-system-accuracy-all-models}), suggesting that LLMs struggle to interpret fused ring systems from IUPAC names. 
Additionally, in terms of reasoning tokens used, MolJSON was \(\sim 1.8\times\) as efficient as SMILES when completing this task (1,021 vs. 1,854 average output tokens, respectively, Figure \ref{fig:gpt5-shortest-path}b).
These results show that the input representation affects graph reasoning performance, with MolJSON the most reliable and reasoning-token-efficient representation among those tested. 
The increased token utilisation for SMILES and IUPAC is consistent with the LLM having to reconstruct the molecular graph before performing a graph search, and the lower accuracy suggests this parsing step is error-prone.

\subsection{MolJSON Substantially Improves Accuracy in Constrained Molecular Generation}
\label{sec:constrained-generation}

\begin{figure}[t]
  \centering
  \includegraphics[width=0.5\linewidth]{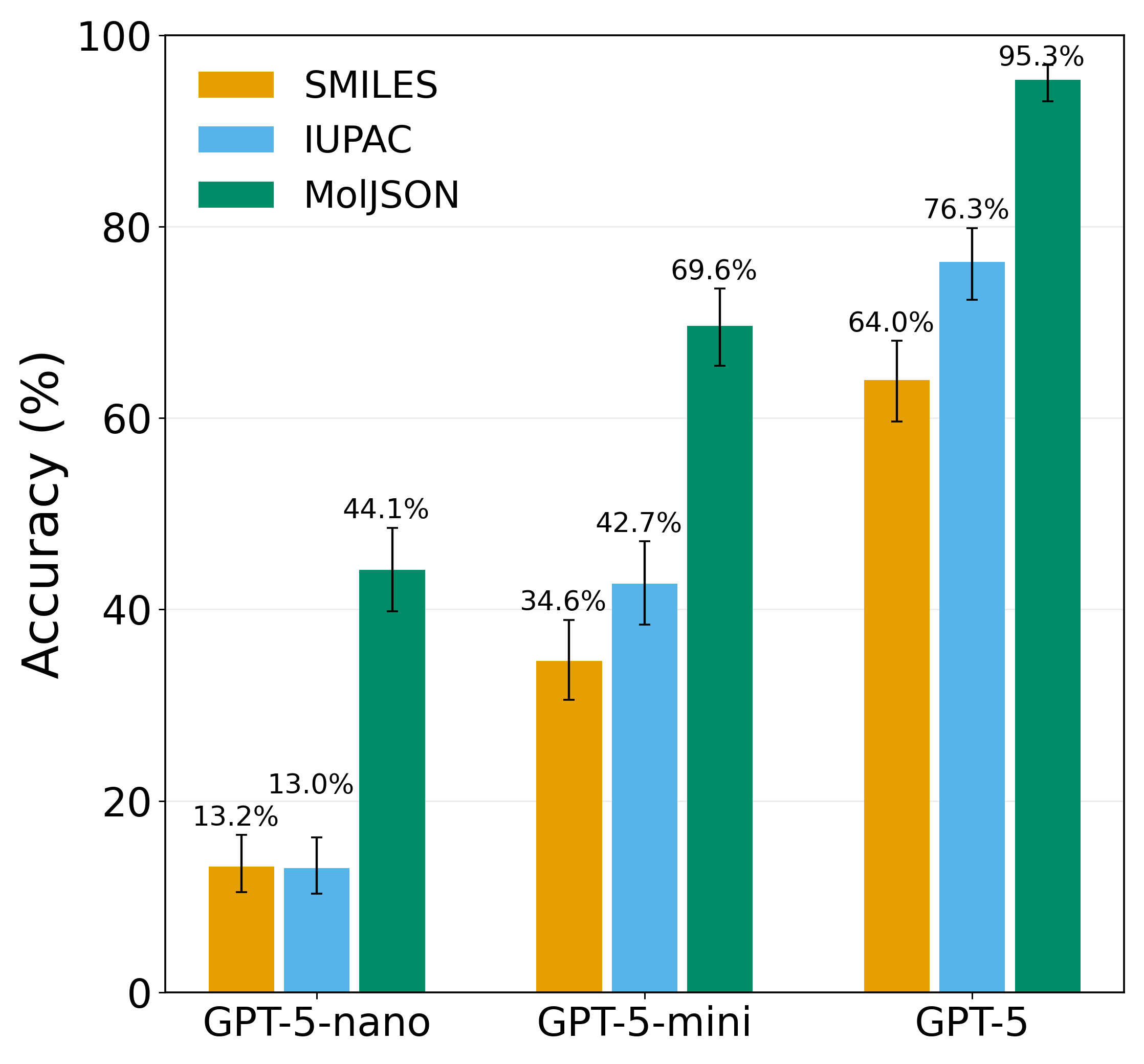}
  \caption{Constrained generation task performance. Models were prompted to generate a molecule consistent with a set of constraints. $494$ questions were generated for each molecular representation. Error bars represent 95\% Wilson confidence intervals.}
  \label{fig:constrained-generation-aggregate}
\end{figure}

To isolate the impact of the output representation, we created a set of constrained molecular generation tasks. 
We evaluated 494 constraint sets for each of SMILES, IUPAC, and MolJSON, yielding a total of 1,482 questions. 
Across GPT-5-nano, GPT-5-mini, and GPT-5, MolJSON was consistently the most reliable output format (Figure \ref{fig:constrained-generation-aggregate}).
For example, GPT-5 achieved 95.3\% accuracy with MolJSON, compared with 64.0\% for SMILES and 76.3\% for IUPAC. 

The performance increase for MolJSON depended on the LLM and on molecular topology (Appendix \ref{app:constrained-mol-gen-result} Table \ref{tab:constrained-generation-accuracy-by-model-format}). 
For GPT-5-mini and GPT-5, the largest gains were observed for tasks involving two rings; GPT-5 achieved 92\% accuracy when generating fused ring systems with MolJSON, compared with 42\% for SMILES and 56\% with IUPAC names. 
The improvement in performance of GPT-5-nano was primarily driven by increased accuracy on the acyclic and monocyclic subsets. 
Additionally, for molecules with two rings, GPT-5-nano achieved less than 1\% accuracy when outputting molecules in SMILES and IUPAC formats, but 13.3\% accuracy when using MolJSON (Appendix \ref{app:constrained-mol-gen-result} Table \ref{tab:constrained-generation-accuracy-by-model-format}).
These results show that the output representation significantly affects the ability of an LLM to write a molecule, with MolJSON being the most reliable format tested. 

\subsection{Performance Degrades with Increasing Heavy-Atom Count and Ring Complexity}
\label{sec:error-analysis}

\begin{figure}[t]
  \centering
  \includegraphics[width=\linewidth]{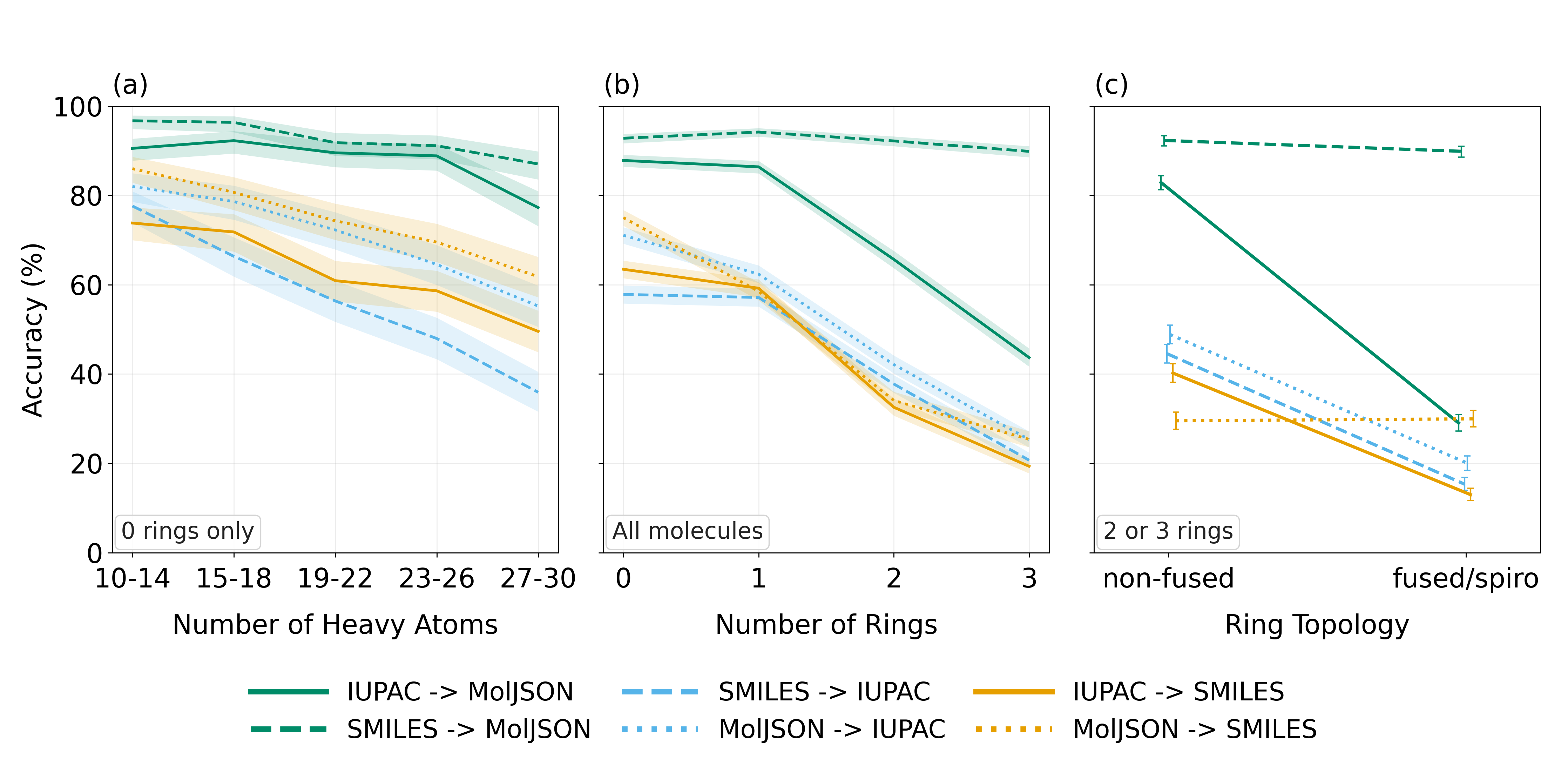}
  \caption{Error modes identified for translation tasks with GPT-5. (a) Translation accuracy against number of heavy atoms for the subset of molecules with zero rings (b) translation accuracy against ring count across all molecules (c) translation accuracy comparing non-fused and fused or spiro ring systems, using the subset of molecules with two or three rings. Shaded region and error bars represent 95\% Wilson confidence intervals.}
  \label{fig:gpt5-error-modes-panel}
\end{figure}

The results in Sections \ref{sec:general-translation} to \ref{sec:constrained-generation} showed that LLM performance varied markedly depending on the molecular representation, and suggested that each format is associated with distinct failure modes.
To investigate these error modes further, we systematically analysed the translation responses from Section \ref{sec:translation-smiles-iupac-moljson}.
In general, error modes corresponded to molecular size, ring count, and ring topology (Figure \ref{fig:gpt5-error-modes-panel}).
We focus the following analysis on GPT-5; similar trends were observed for GPT-5-mini and GPT-5-nano except when stated (Appendix \ref{sec:additional_translation_error_analysis} Figure \ref{fig:gpt5mini-nano-error-modes-panel}). 
Full results for each model stratified by ring and heavy-atom count can be found in Appendix \ref{sec:additional_translation_error_analysis} Figure \ref{fig:translation-ring-and-heavy-atom-count}. 
We restrict comparisons to specific subsets of molecules to minimise overlap between error modes. 

For molecules with zero rings, accuracy decreased with increasing heavy atom count (Figure \ref{fig:gpt5-error-modes-panel}a). 
The MolJSON output was most robust to this trend, indicating GPT-5 can reliably parse SMILES and IUPAC inputs into a molecular graph even for larger molecules.
In comparison, the SMILES and IUPAC output performance dropped significantly with larger size.
Together, for molecules with zero rings, these results are consistent with errors occurring during SMILES and IUPAC generation rather than input parsing. 

Accuracy decreased with increasing ring count across most translation directions (Figure \ref{fig:gpt5-error-modes-panel}b).
The SMILES to MolJSON result for GPT-5 was a notable exception, with performance staying consistent across all ring counts, whereas for GPT-5-mini and GPT-5-nano accuracy decreased with ring count (Appendix \ref{sec:additional_translation_error_analysis} Figure \ref{fig:gpt5mini-nano-error-modes-panel}). 
Possible explanations for the higher accuracy of GPT-5 at parsing SMILES strings include architectural differences, such as its larger model size, and training differences, which could include additional specific training on SMILES strings. 

The IUPAC to MolJSON task had similar performance for zero and one ring, then steeply declined for two and three rings (Figure \ref{fig:gpt5-error-modes-panel}b), 
which can be attributed to the presence of fused ring systems within the molecule (Figure \ref{fig:gpt5-error-modes-panel}c).
Similarly, GPT-5 made more errors writing the IUPAC names of molecules with fused ring systems (Figure \ref{fig:gpt5-error-modes-panel}c).
Consistent with the shortest-path and constrained generation task findings, this indicates fused ring systems are a general error mode when using IUPAC names as a molecular representation with LLMs. 

A similar trend across ring counts was observed for the IUPAC to SMILES task, but with consistently lower accuracy than for IUPAC to MolJSON (Figure \ref{fig:gpt5-error-modes-panel}b).
This suggests that, in addition to errors arising from interpreting IUPAC names, further failure modes occurred during SMILES generation.
We were unable to identify specific structural motifs that explained these SMILES output errors; however, across the GPT-5 translation responses, SMILES outputs produced the greatest proportion of invalid molecules, suggesting the SMILES syntax itself as a source of error (Appendix \ref{sec:additional_translation_error_analysis} Figure \ref{fig:translation-error-stage}).

Our analysis suggests MolJSON is more robust to size- and topology-dependent failure modes than SMILES and IUPAC. 
This is likely due to MolJSON representing molecules as explicit molecular graphs, while avoiding the traversal constraints of SMILES strings and the rule-based nomenclature mapping of IUPAC names.

\section{Conclusion}
Our results show that the choice of molecular representation materially impacts the ability of LLMs to interpret and output molecular structures, and the performance of LLMs in chemistry could currently be limited by the use of existing cheminformatics formats. 
Across all tasks, our structured JSON format, MolJSON, consistently outperformed all other tested molecular representations as both an input and an output format.
For example, on translation tasks where the model had to convert an IUPAC name into another format, GPT-5 achieved 43.7\% accuracy when outputting a SMILES string compared with 71.0\% using MolJSON as the output format. 
This advantage was observed even though SMILES strings and IUPAC names are ubiquitous in public chemical databases and were likely well represented during model training, whereas the evaluated models were not explicitly trained to use MolJSON.
This suggests that molecular graph representations that explicitly encode atoms and bonds are intrinsically better aligned with current LLMs than more standard serialised formats. 
We therefore suggest that structured representations of molecular graphs, such as MolJSON, should be used when developing LLM-based systems for chemistry.

A likely reason for the improved performance of MolJSON is that it provides a more explicit and less restrictive representation of molecular graphs than other molecular formats.
Traversal-based encodings, such as SMILES, and rule-based encodings, such as IUPAC, require molecular structures be expressed within strict syntactic and semantic constraints. 
On input tasks, these formats require the model to infer the graph structure before it can reason over the molecule, and, on output tasks, they require the intended molecule to be serialised into a strict form. 
MolJSON avoids these challenges by presenting the atoms and bonds explicitly and allowing the LLM to output the structure directly.
The systematic error modes observed for SMILES and IUPAC relating to molecular size and ring topology support the view that these formats are intrinsically more challenging for LLMs.
In each case, MolJSON was more robust to these error modes.

Our benchmark was designed to isolate the ability of LLMs to operate on molecular structures across different representations, rather than to assess LLM performance on applied chemistry tasks. 
This allowed us to determine specific strengths and weaknesses of each format while controlling for confounding effects from chemical knowledge. 
Although we did not evaluate the models on more complex applied chemistry problems, we believe our tasks assess prerequisite skills required for downstream applications; for example, the constrained generation task mirrors the task of structure elucidation. 
We expect the foundational limitations identified for each format to persist in downstream chemistry tasks. 

The MolJSON schema used in this work should be viewed as a starting point for LLM-oriented molecular representations rather than as a definitive format.
There are multiple ways in which our schema could be modified and improved.
For example, we represented bonds using ``source'' and ``target'' keys; however, a two-item list may be more appropriate given the undirected nature of chemical bonds.
Similarly, explicit protons could be represented directly within the atoms list rather than through a dedicated key for aromatic nitrogens.
Additionally, this format could be extended to support other molecular features we did not consider, such as stereochemistry and atomic coordinates. 

Our study represents an initial exploration of molecular representations designed specifically for LLMs. 
Our results demonstrate that the choice of molecular representation is an important consideration when developing LLM-based systems for chemistry.
The improved performance of MolJSON over established formats such as SMILES and IUPAC, despite LLMs having been explicitly trained on those representations, indicates the promise of molecular representations tailored to LLMs.
We hope this motivates the broader exploration of molecular representations designed specifically for LLMs.

\section*{Software and Data Availability}
The MolJSON schema and scripts are available at \mbox{\url{https://github.com/oxpig/MolJSON}}, and the task prompts, model responses, and analysis scripts are available at \mbox{\url{https://github.com/oxpig/MolJSON-data}}.

\section*{Acknowledgments}
We acknowledge OpenAI for providing the API credits used to run the experiments in this study.
N.T.R. is supported by UK Research and Innovation (UKRI) through an EPSRC-funded PhD studentship (project reference: 2928891).

\bibliographystyle{plainnat6doiElseURL.bst}
\bibliography{references}

\appendix

\clearpage
\section{Additional Methods Details}

\subsection{Shortest-Path Molecule Sampling}
\label{sec:shortest-path-length-counts}

\begin{table}[H]
\centering
\caption{Distribution of molecules sampled in the shortest-path questions. ``Non-fused'' refers to the counts of molecules which do not have any fused ring systems, and ``fused'' refers to molecules that contain at least one fused ring system.}
\small
\setlength{\tabcolsep}{4pt}
\begin{tabular}{rccccccccccccccccc}
\toprule
Length & 2 & 3 & 4 & 5 & 6 & 7 & 8 & 9 & 10 & 11 & 12 & 13 & 14 & 15 & 16 & 17 & 18 \\
\midrule
Total & 200 & 200 & 200 & 200 & 200 & 200 & 200 & 200 & 200 & 200 & 200 & 200 & 200 & 151 & 79 & 52 & 37 \\
Non-fused & 131 & 104 & 120 & 122 & 62 & 98 & 97 & 100 & 103 & 104 & 100 & 91 & 134 & 79 & 60 & 33 & 28 \\
Fused & 69 & 96 & 80 & 78 & 138 & 102 & 103 & 100 & 97 & 96 & 100 & 109 & 66 & 72 & 19 & 19 & 9 \\
\bottomrule
\end{tabular}
\label{tab:shortest-path-length-counts}
\end{table}

\subsection{Constrained Molecular Generation Task Design}
\label{app:constrained-mol-gen-methods}

\begin{figure}[H]
  \centering
  \includegraphics[width=\linewidth]{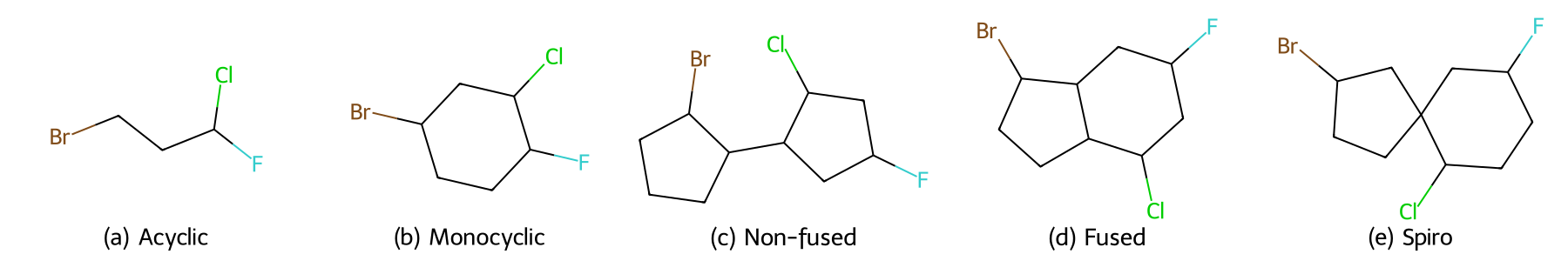}
  \caption{Example witness molecules used when creating the constrained molecular generation questions. These molecules were used to ensure at least one valid solution existed for each question.}
  \label{fig:example-witness-molecules}
\end{figure}

The constrained molecular generation tasks were constructed by defining a set of constraints, and prompting the LLM to generate a molecule consistent with them. 
Any generated molecule consistent with all constraints was accepted as correct.
Each task was generated with a corresponding ``witness'' molecule to ensure that at least one valid solution existed (Appendix \ref{app:constrained-mol-gen-methods} Figure \ref{fig:example-witness-molecules}).
Five subsets of tasks were created: acyclic, monocyclic, non-fused, fused, and spiro (Appendix \ref{app:constrained-mol-gen-methods} Table \ref{tab:constrained-generation-subsets}).
All tasks included a shortest-path constraint requiring the molecule to contain one fluorine, one chlorine, and one bromine atom, with specific shortest-path distances between each pair.
Tasks involving ring generation additionally included constraints on (1) the number of rings, (2) ring sizes, (3) ring topology (non-fused, fused, or spiro), and (4) required each halogen atom be directly bonded to a ring atom. 
Molecules were enumerated for each subset to identify valid constraint sets (halogen permutations were treated as equivalent). 
Since enumeration favors larger molecules, the final task set was constructed by uniformly sampling (without replacement) up to 100 tasks across structural strata (ring sizes or total halogen path lengths). An example prompt is shown in Appendix \ref{app:constrained-mol-gen-methods} Figure \ref{fig:example-constrained-generation-prompt}. 

\begin{table}[H]
\centering
\small
\caption{Sampling ranges used to construct the constrained molecular generation task subsets.}
\label{tab:constrained-generation-subsets}
\begin{tabular}{lcccc}
\toprule
Subset & N tasks & Topology & Ring sizes & Halogen shortest paths \\
\midrule
Acyclic    & 100 & 0 rings            & none     & 2 to 13 \\
Monocyclic & 100 & 1 ring             & 3 to 27  & 3 to 11 \\
Non-fused  & 100 & 2 rings, separate  & 3 to 7   & 3 to 9  \\
Fused      & 94  & 2 rings, fused     & 3 to 7   & 3 to 8  \\
Spiro      & 100 & 2 rings, spiro     & 3 to 7   & 3 to 8  \\
\bottomrule
\end{tabular}
\end{table}

\begin{figure}[H]
  \centering
  \begin{tcolorbox}[
    colback=white,
    colframe=black,
    boxrule=0.5pt,
    arc=0pt,
    left=4pt,
    right=4pt,
    top=4pt,
    bottom=4pt
  ]
\begin{lstlisting}[
  basicstyle=\ttfamily\footnotesize,
  frame=none,
  breaklines=true,
  breakatwhitespace=false,
  showstringspaces=false
]
Generate one valid molecule that satisfies all constraints below.
- Connectivity: exactly 1 connected component (single connected molecule).
- Number of F atoms: exactly 1.
- Number of Cl atoms: exactly 1.
- Number of Br atoms: exactly 1.
- Shortest path between F and Cl: exactly 5 bonds (count all bonds on the path, including the bond to each halogen).
- Shortest path between F and Br: exactly 6 bonds (count all bonds on the path, including the bond to each halogen).
- Shortest path between Cl and Br: exactly 5 bonds (count all bonds on the path, including the bond to each halogen).
- Number of rings: exactly 2.
- Ring sizes: exactly one 5-membered ring and one 6-membered ring.
- Ring topology: exactly one spiro center.
- Halogen placement: each halogen must be directly bonded to a ring atom.
Return only the molecule in the requested output format.
\end{lstlisting}

\centering
  \end{tcolorbox}
  \caption{Example prompt for a constrained molecular generation task. For these questions the output format was specified in the JSON schema, rather than being explicit within the prompt. A valid solution to this question is the molecule in Appendix \ref{app:constrained-mol-gen-methods} Figure \ref{fig:example-witness-molecules}e.}
  \label{fig:example-constrained-generation-prompt}
\end{figure}

\subsection{MolJSON Schema}
\label{app:schema-full}

\begin{figure}[H]
  \centering
  \begin{tcolorbox}[
    colback=white,
    colframe=black,
    boxrule=0.5pt,
    arc=0pt,
    left=6pt,
    right=6pt,
    top=6pt,
    bottom=6pt
  ]
\begin{lstlisting}[
  basicstyle=\ttfamily\normalsize,
  frame=none,
  breaklines=true,
  breakatwhitespace=false,
  showstringspaces=false
]
{'type': 'object',
 'additionalProperties': False,
 'required': ['atoms', 'bonds', 'charges', 'aromatic_n_h'],
 'properties': {'atoms': {'type': 'array',
   'items': {'type': 'object',
    'additionalProperties': False,
    'properties': {'id': {'type': 'string', 'description': 'Unique atom id.'},
     'element': {'type': 'string',
      'enum': ['*','H','He','Li','Be','B','C','N','O','F','Ne','Na','Mg','Al','Si','P','S','Cl','Ar','K','Ca','Sc','Ti','V','Cr','Mn','Fe','Co','Ni','Cu','Zn','Ga','Ge','As','Se','Br','Kr','Rb','Sr','Y','Zr','Nb','Mo','Tc','Ru','Rh','Pd','Ag','Cd','In','Sn','Sb','Te','I','Xe','Cs','Ba','La','Ce','Pr','Nd','Pm','Sm','Eu','Gd','Tb','Dy','Ho','Er','Tm','Yb','Lu','Hf','Ta','W','Re','Os','Ir','Pt','Au','Hg','Tl','Pb','Bi','Po','At','Rn','Fr','Ra','Ac','Th','Pa','U','Np','Pu','Am','Cm','Bk','Cf','Es','Fm','Md','No','Lr','Rf','Db','Sg','Bh','Hs','Mt','Ds','Rg','Cn','Nh','Fl','Mc','Lv','Ts','Og'],
      'description': "Element symbol like 'C' or 'Cl', or '*' dummy atom."}},
    'required': ['id', 'element']}},
  'bonds': {'type': 'array',
   'items': {'type': 'object',
    'additionalProperties': False,
    'properties': {'source': {'type': 'string'},
     'target': {'type': 'string'},
     'order': {'type': 'number',
      'enum': [0, 1, 1.5, 2, 3],
      'description': 'Bond order. Aromatic bonds are 1.5. ZERO bonds are 0.'}},
    'required': ['source', 'target', 'order']}},
  'charges': {'type': ['array', 'null'],
   'description': 'Sparse list of NON-ZERO formal charges. Null means none.',
   'items': {'type': 'object',
    'additionalProperties': False,
    'properties': {'atom_id': {'type': 'string'},
     'formal_charge': {'type': 'integer', 'minimum': -5, 'maximum': 5}},
    'required': ['atom_id', 'formal_charge']}},
  'aromatic_n_h': {'type': ['array', 'null'],
   'description': 'Sparse list of aromatic nitrogens with explicit hydrogen count. Null means none.',
   'items': {'type': 'object',
    'additionalProperties': False,
    'properties': {'atom_id': {'type': 'string'},
     'hcount': {'type': 'integer', 'minimum': 1, 'maximum': 2}},
    'required': ['atom_id', 'hcount']}}}}
\end{lstlisting}
  \end{tcolorbox}
  \caption{The structured JSON schema which was used when submitting prompts to the OpenAI API.}
  \label{fig:schema-full}
\end{figure}

\clearpage

\section{Additional Results and Analysis}

\subsection{Analysis of MOL V2000, SELFIES, and InChI Failure Modes}
\label{app:mol-v2000}

The translation accuracy of MOL V2000 to MolJSON for GPT-5-mini was 81.7\%. In 100\% of outputs the MolJSON atom identifiers took the form \texttt{1,2,3} or \texttt{a1,a2,a3}. 
In all responses which were correct, the integer numbering corresponded to the numbering used within the MOL V2000 file to index atoms and bonds. 
This indicates the MOL V2000 was likely trivially reformatted into the MolJSON format, rather than the molecular graph being explicitly parsed from the MOL V2000 files. 
The MOL V2000 to SMILES accuracy was 15.0\%, and MOL V2000 to IUPAC accuracy was 10.7\%, both of which are lower than the equivalent tasks using SMILES, IUPAC, or MolJSON inputs. 
These results indicate that the GPT-5-mini model can sometimes parse the molecular graph from the MOL V2000 format, however this format is a poorer input representation than SMILES, IUPAC, and MolJSON. 

The MOL V2000 performance was asymmetric, with MolJSON to MOL V2000 having an accuracy of 22.1\%. 
A frequent error mode was in the writing of the header of the MOL V2000 file, specifically the ``counts line''. 
By using an algorithmic parser which uses heuristics to identify the atom and bond blocks and then reconstructs the molecule directly, the output accuracy of MOL V2000 could be improved further (Appendix \ref{app:mol-v2000} Table \ref{tab:v2000_rescue_before_after_by_input}). Even with this parsing script, the accuracy of MOL V2000 outputs was still lower than SMILES and IUPAC outputs. These findings suggest the strict formatting requirements of MOL V2000 make this an unreliable format for LLMs to output, and as such was not considered further. 

\begin{table}[H]
\centering
\small
\caption{GPT-5-mini translation accuracy to the MOL V2000 format, with and without algorithmic parsing of the output. A common error mode was that GPT-5-mini wrote the ``counts line'' on the incorrect line in the file. The output accuracy of MOL V2000 could be increased with a script that uses heuristics to identify the atom and bond blocks, then manually reconstructs the molecule.}
\begin{tabular}{lcc}
\toprule
Input format & Before parser (\%) & After parser (\%) \\
\midrule
SMILES   & 8.6 & 15.0 \\
IUPAC    & 4.0 & 12.6 \\
MolJSON  & 22.1 & 41.2 \\
SELFIES  & 0.7 & 1.0 \\
InChI    & 0.2 & 0.7 \\
\bottomrule
\end{tabular}
\label{tab:v2000_rescue_before_after_by_input}
\end{table}

Across all translation questions, 2100 involved outputting a SELFIES string. 
Of all GPT-5-mini responses, 1033 were an empty string (\texttt{\{"selfies":""}\}) indicating the model failed or refused to respond. 
Of the non-empty strings, 314 did not contain a `\texttt{[}' or `\texttt{]}' character, indicating the model did not attempt to write a SELFIES string; these responses were model refusals, for example ``CONVERSION\_NOT\_POSSIBLE''. 
There were 753 responses that contained square brackets, indicating an attempt was made to write the SELFIES string. 
Of these, 394 responses were valid SELFIES strings when using the standard SELFIES decoder; this number increased to 512 when setting the \texttt{compatible=True} flag, which allows for legacy SELFIES syntax to be decoded. 
The invalid SELFIES strings corresponded to malformed strings, for example \texttt{[C][C][C][c][c][c][n]} contains invalid tokens.
Across the valid SELFIES strings, only 13 responses matched the expected ground truth structure; all of these correct structures were for linear molecules with zero branches and zero rings. 

Across all translation questions, 2100 involved outputting an InChI string. 
Of all GPT-5-mini responses, 1369 were an empty string (\texttt{\{"inchi":""}\}). 
There were 695 responses that were non-empty text, but invalid InChI strings; 125 of these were just the prefix \texttt{InChI=1S/}, 106 contained refusal prose, and the others were malformed InChI responses. 
There were only 36 valid InChI strings generated, 30 of which were methane, and none of these corresponded to the correct molecule.

\clearpage
\subsection{Translation Results for Claude Haiku 4.5}
\label{sec:haiku-4-5-translation}

\begin{figure}[H]
  \centering
  \includegraphics[width=0.6\linewidth]{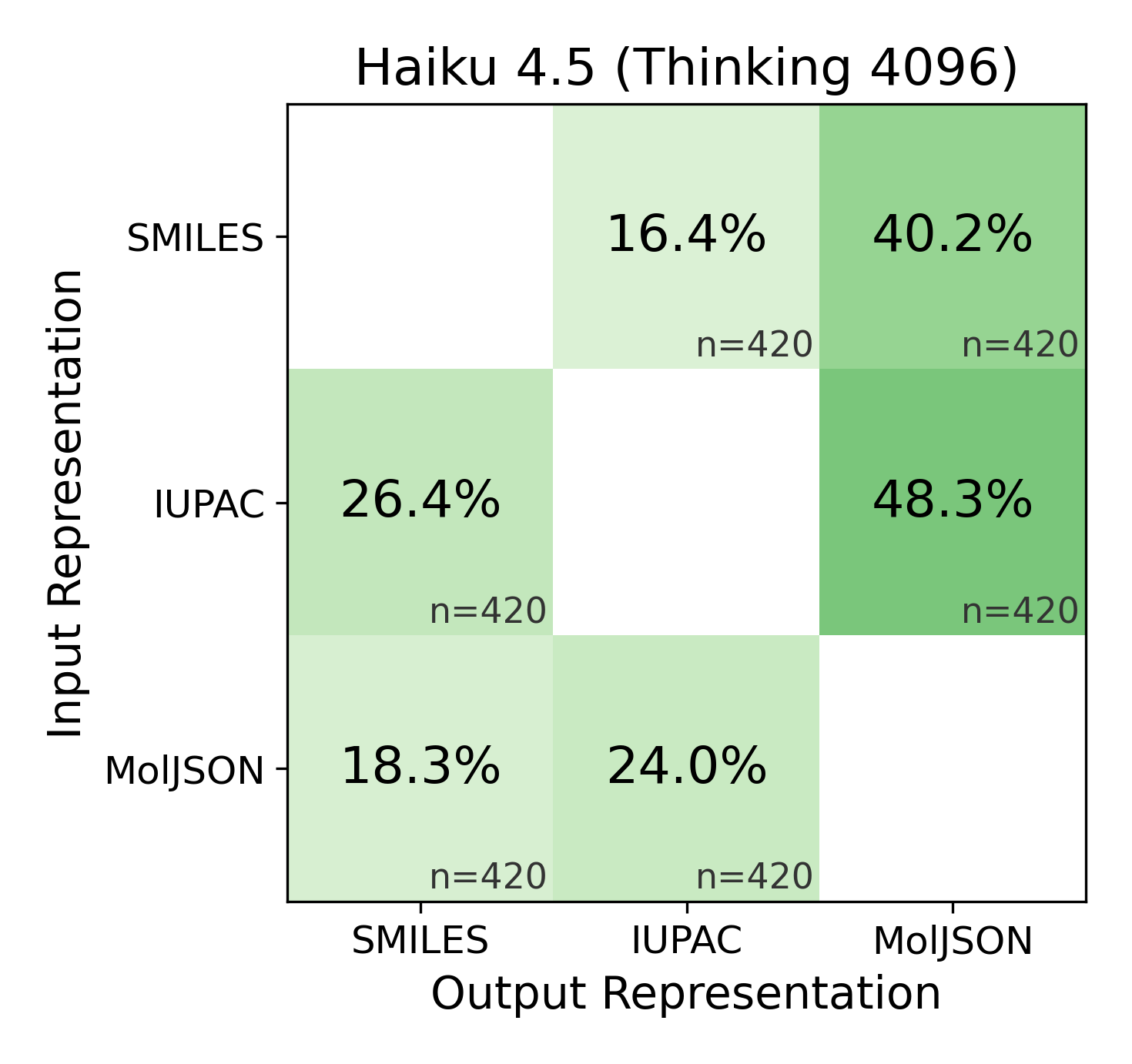}
  \caption{Accuracy of Claude Haiku 4.5 on pairwise translation tasks between SMILES, IUPAC, and MolJSON representations. Questions were selected from the small translation set of 420 molecules (2520 questions total). The model was run with a thinking budget of 4096 tokens, and maximum output tokens of 16,384. The model was run using JSON output mode with a slightly modified MolJSON schema where numeric range fields were replaced with integer enumerations for Anthropic API compatibility. }
  \label{fig:haiku-4-5-translation}
\end{figure}

\clearpage
\subsection{Shortest-Path Task Results}
\label{sec:shortest_path_task_results}

\begin{table}[H]
\centering
\caption{Aggregate shortest-path accuracy for each model and input representation. 
Accuracy is reported as the mean proportion correct, with 95\% confidence interval half-widths computed as $1.96 \times$ the standard error for the binomial proportion. Output tokens are reported as the mean number of output tokens, with 95\% confidence interval half-widths computed as $1.96 \times$ the standard error of the mean.}
\label{tab:shortest_path_by_input}
\begin{tabular}{llccc}
\toprule
Input & Metric & GPT-5 & GPT-5-mini & GPT-5-nano \\
\midrule
SMILES & Accuracy (\%) & 92.2 $\pm$ 1.0 & 53.3 $\pm$ 1.8 & 42.9 $\pm$ 1.8 \\
       & Output Tokens & 1854 $\pm$ 37  & 1263 $\pm$ 25 & 1687 $\pm$ 31 \\
\midrule
IUPAC  & Accuracy (\%) & 82.7 $\pm$ 1.4 & 63.4 $\pm$ 1.7 & 42.7 $\pm$ 1.8 \\
       & Output Tokens & 1587 $\pm$ 35  & 881 $\pm$ 18  & 1351 $\pm$ 25 \\
\midrule
MolJSON & Accuracy (\%) & \textbf{98.5 $\pm$ 0.5} & \textbf{86.6 $\pm$ 1.2} & \textbf{67.7 $\pm$ 1.7} \\
        & Output Tokens & \textbf{1021 $\pm$ 22}  & \textbf{615 $\pm$ 13}   & \textbf{1150 $\pm$ 21} \\
\bottomrule
\end{tabular}
\end{table}

\begin{figure}[H]
  \centering
  \includegraphics[width=0.95\linewidth]{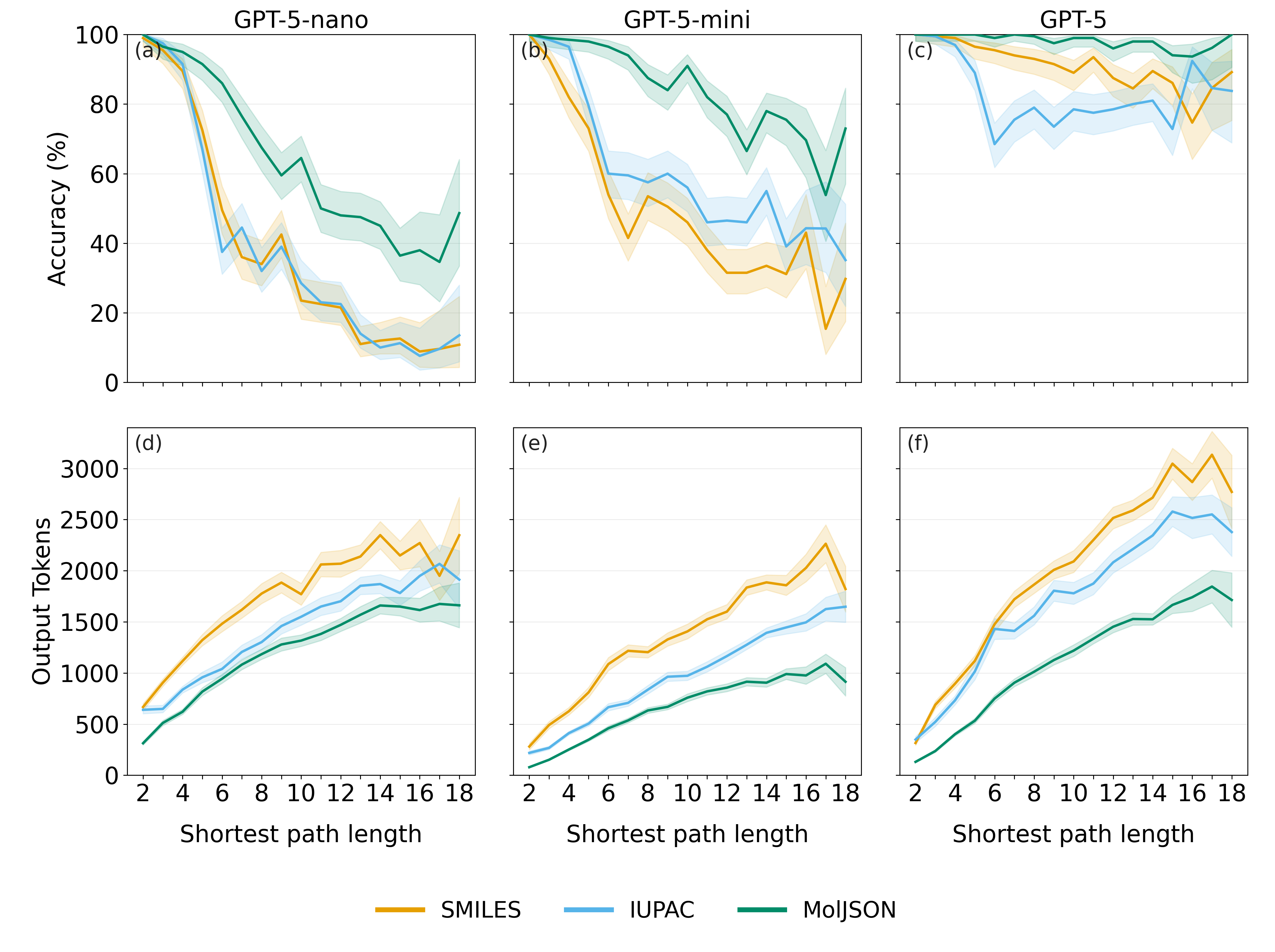}
  \caption{Performance of GPT-5-nano, GPT-5-mini, and GPT-5 on shortest-path tasks by path length. Models were prompted to determine the number of bonds on the shortest path between two halogen atoms. (a,b,c) show accuracy by path length, (d,e,f) show average output tokens used per question by path lengths. Shaded regions show 95\% confidence interval.}
  \label{fig:shortest-path-length}
\end{figure}

\begin{figure}[H]
  \centering
  \includegraphics[width=0.8\linewidth]{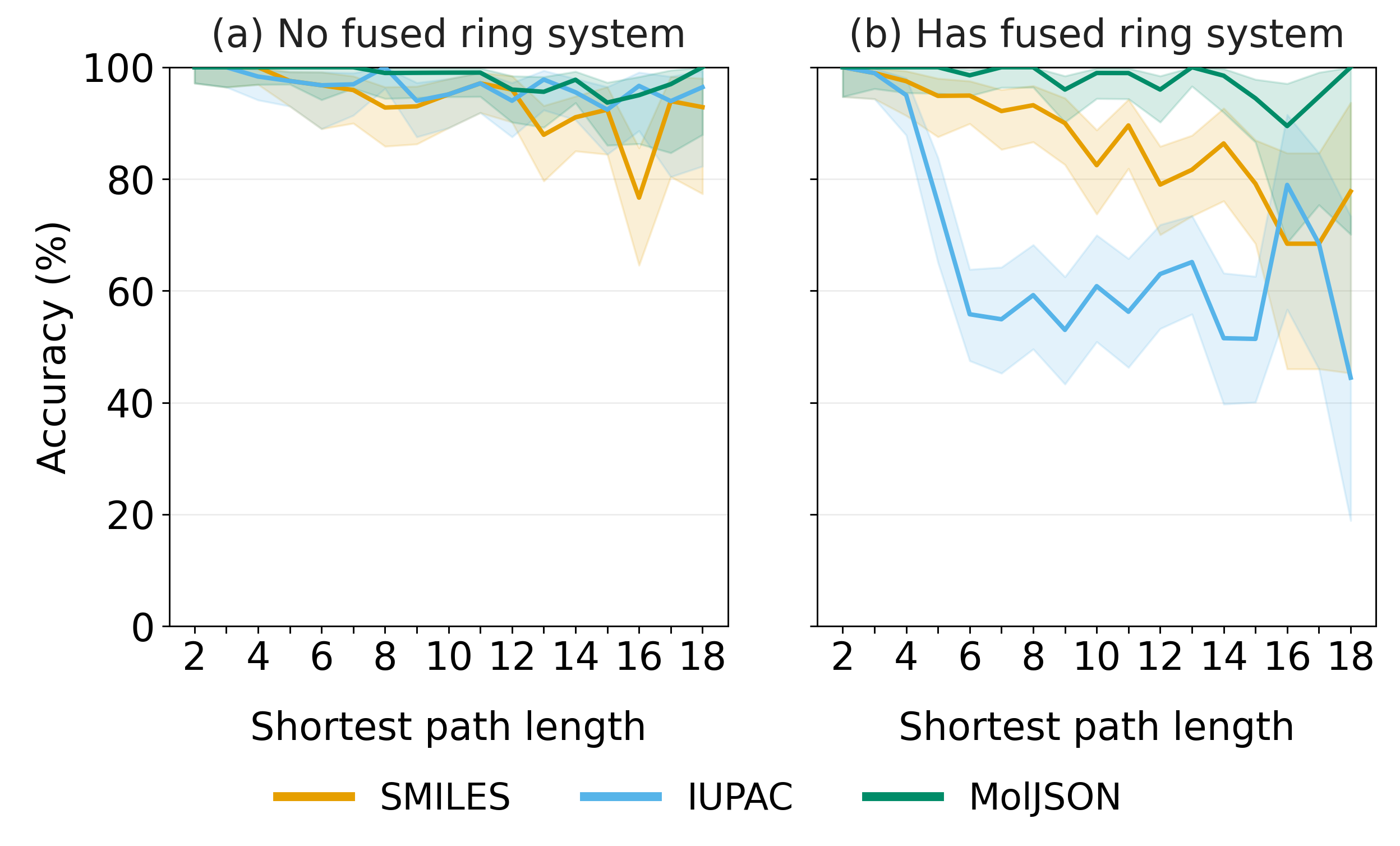}
  \caption{Accuracy of GPT-5 on shortest-path questions, where (a) shows the subset of molecules with no fused rings, and (b) shows molecules that contain a fused ring system.}
  \label{fig:shortest-path-length-split-fused}
\end{figure}

\begin{table}[H]
\centering
\caption{Aggregate accuracy of GPT-5-nano, GPT-5-mini, and GPT-5 on shortest-path tasks for each input representation, grouped by whether the question molecule contained a fused ring system. The total non-fused subset size of 1,566 molecules, and the fused-ring subset was 1,353 molecules.} 
\begin{tabular}{llcc}
\toprule
Model & Input format & No fused ring system & Has fused ring system \\
\midrule
GPT-5 & SMILES & 94.8\% & 89.3\% \\
GPT-5 & IUPAC & 96.9\% & 66.2\% \\
GPT-5 & MolJSON & 98.5\% & 98.5\% \\
\addlinespace
GPT-5-mini & SMILES & 62.6\% & 42.5\% \\
GPT-5-mini & IUPAC & 81.9\% & 42.1\% \\
GPT-5-mini & MolJSON & 87.4\% & 85.7\% \\
\addlinespace
GPT-5-nano & SMILES & 48.3\% & 36.7\% \\
GPT-5-nano & IUPAC & 51.2\% & 32.9\% \\
GPT-5-nano & MolJSON & 71.9\% & 62.8\% \\
\bottomrule
\end{tabular}
\label{tab:shortest-path-fused-system-accuracy-all-models}
\end{table}

\clearpage
\subsection{Constrained Molecular Generation Task Results}
\label{app:constrained-mol-gen-result}

\begin{table}[H]
\centering
\small
\caption{Accuracy (\%) of GPT-5-nano, GPT-5-mini, and GPT-5 on constrained molecular generation tasks, grouped by question subset.}
\label{tab:constrained-generation-accuracy-by-model-format}
\begin{tabular}{llccc}
\toprule
Model & Subset & SMILES & IUPAC & MolJSON \\
\midrule
GPT-5
& No Rings (n=100)         & 89.0 & \textbf{100.0} & 99.0 \\
& 1 Ring (n=100)           & 67.0 & 97.0  & \textbf{100.0} \\
& 2 Rings Separate (n=100) & 64.0 & 70.0  & \textbf{96.0} \\
& 2 Rings Fused (n=94)     & 41.5 & 56.4  & \textbf{91.5} \\
& 2 Rings Spiro (n=100)    & 57.0 & 57.0  & \textbf{90.0} \\
\midrule
GPT-5-mini
& No Rings (n=100)         & 81.0 & 68.0 & \textbf{95.0} \\
& 1 Ring (n=100)           & 52.0 & 92.0 & \textbf{95.0} \\
& 2 Rings Separate (n=100) & 16.0 & 30.0 & \textbf{49.0} \\
& 2 Rings Fused (n=94)     & 18.1 & 11.7 & \textbf{59.6} \\
& 2 Rings Spiro (n=100)    & 5.0  & 10.0 & \textbf{49.0} \\
\midrule
GPT-5-nano
& No Rings (n=100)         & 38.0 & 33.0 & \textbf{84.0} \\
& 1 Ring (n=100)           & 26.0 & 29.0 & \textbf{95.0} \\
& 2 Rings Separate (n=100) & 1.0  & 0.0  & \textbf{7.0} \\
& 2 Rings Fused (n=94)     & 0.0  & 1.1  & \textbf{13.8} \\
& 2 Rings Spiro (n=100)    & 0.0  & 1.0  & \textbf{19.0} \\
\bottomrule
\end{tabular}
\end{table}

\clearpage
\subsection{Additional Translation Error Analysis}
\label{sec:additional_translation_error_analysis}

\begin{figure}[H]
  \centering
  \includegraphics[width=0.9\linewidth]{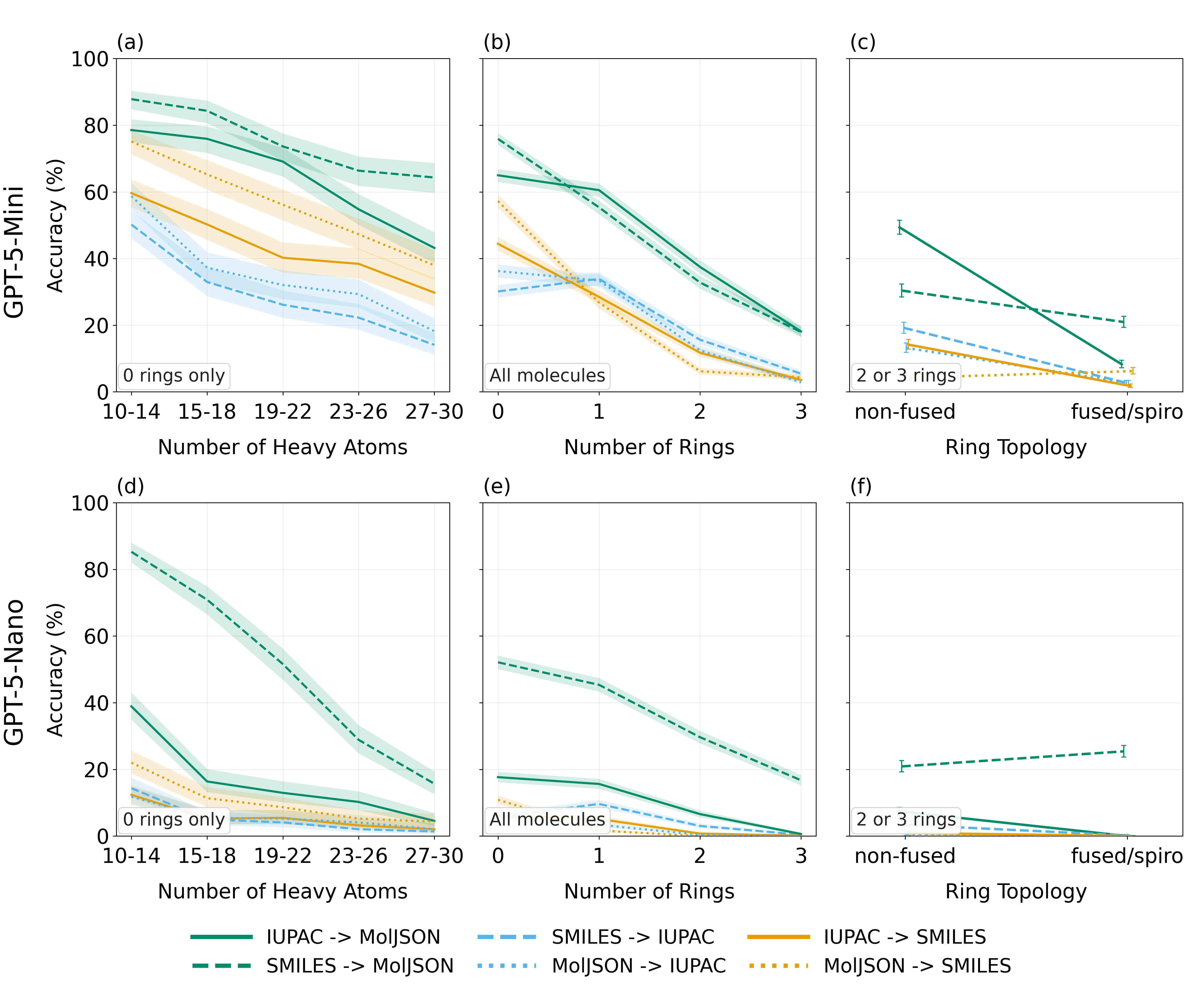}
  \caption{Error modes identified for translation tasks. (a,b,c) shows GPT-5-mini, and (d,e,f) shows GPT-5-nano. (a,d) Translation accuracy against number of heavy atoms for the subset of molecules with zero rings; (b,e) translation accuracy against ring count across all molecules; (c,f) translation accuracy comparing non-fused and fused or spiro ring systems, using the subset of molecules with two or three rings. Shaded region and error bars represent 95\% Wilson confidence intervals.}
  \label{fig:gpt5mini-nano-error-modes-panel}
\end{figure}

\begin{figure}[H]
  \centering
  \includegraphics[width=\linewidth]{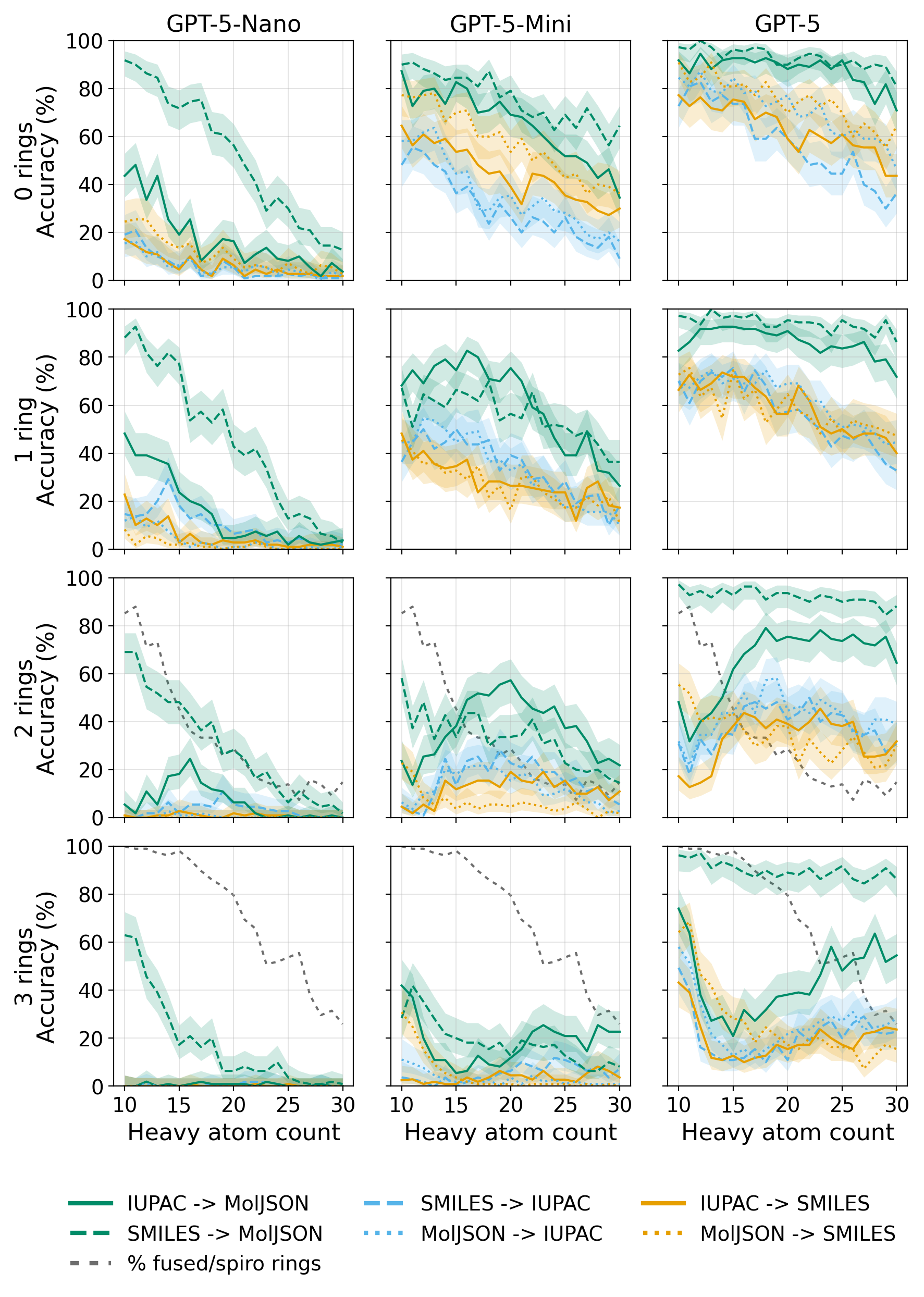}
  \caption{Translation accuracy of GPT-5-nano, GPT-5-mini, and GPT-5 on pairwise translations between SMILES, IUPAC, and MolJSON, using a set of 9,201 unique molecules (55,206 questions total). Each model is shown as a separate column, each row shows results for the specified ring count, and the x-axis of each subplot shows accuracy for each heavy atom count. For subplots with two or three rings, the proportion of these molecules that contain a fused or spiro ring system is plotted as a dotted grey line.}
  \label{fig:translation-ring-and-heavy-atom-count}
\end{figure}

\begin{figure}[H]
  \centering
  \includegraphics[width=0.9\linewidth]{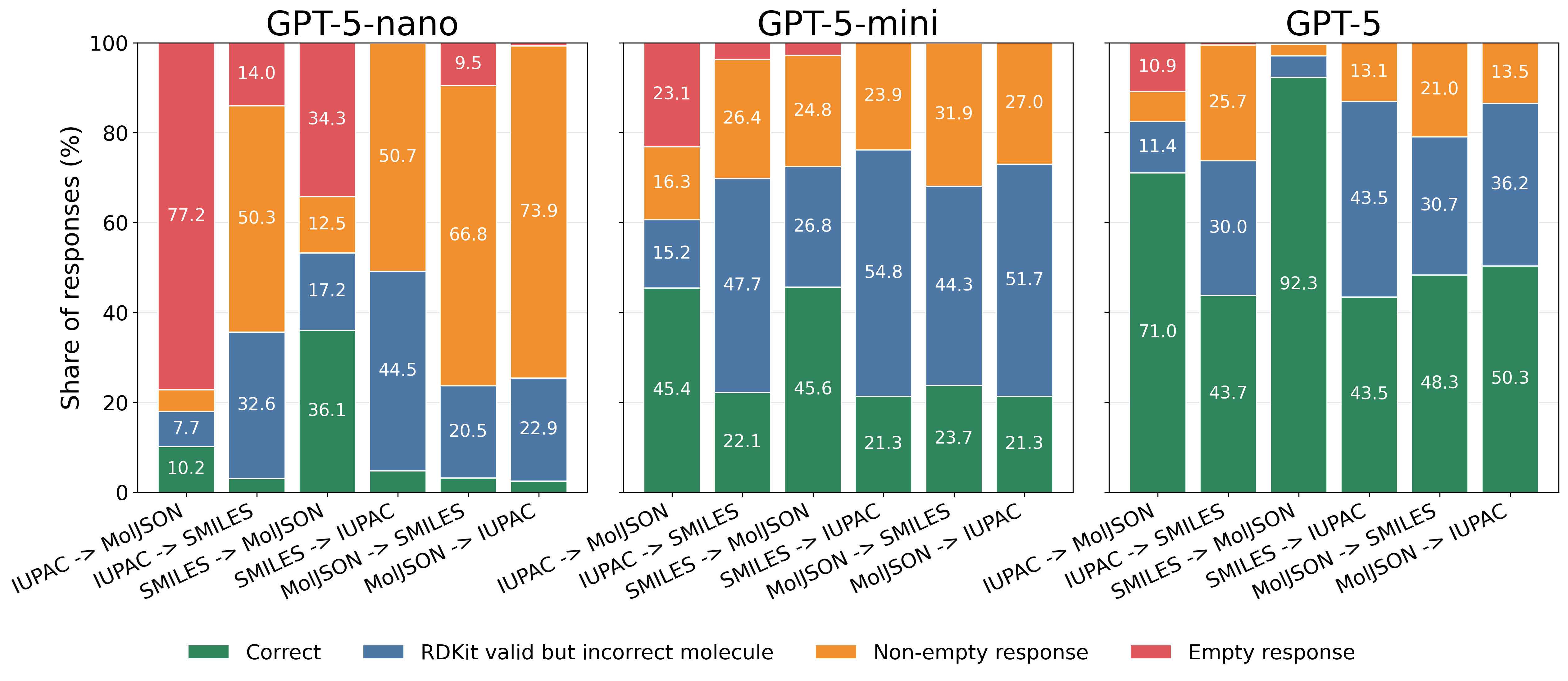}
  \caption{Error stages for translation questions (large set). From bottom to top, (green) correct answer; (blue) RDKit valid output, but wrong molecule; (orange) non-empty response that is not a valid molecule; this includes refusal text and malformed strings; (red) responses that are an empty string. For MolJSON output, responses with zero atoms were considered an empty response.}
  \label{fig:translation-error-stage}
\end{figure}

\section{Example MolJSON Outputs}

\begin{figure}[H]
  \centering
  \begin{tcolorbox}[
    colback=white,
    colframe=black,
    boxrule=0.5pt,
    arc=0pt,
    left=4pt,
    right=4pt,
    top=4pt,
    bottom=4pt
  ]
\begin{lstlisting}[
  basicstyle=\ttfamily\footnotesize,
  frame=none,
  breaklines=true,
  breakatwhitespace=false,
  showstringspaces=false
]
{
  "atoms": [
    {"id": "C_acyl", "element": "C"},
    {"id": "O_carb", "element": "O"},
    {"id": "C_me", "element": "C"},
    {"id": "N1", "element": "N"},
    {"id": "C2", "element": "C"},
    {"id": "C3", "element": "C"},
    {"id": "C4", "element": "C"},
    {"id": "C5", "element": "C"},
    {"id": "O_me", "element": "O"},
    {"id": "C_ome", "element": "C"}
  ],
  "bonds": [
    {"source": "C_acyl", "target": "O_carb", "order": 2},
    {"source": "C_acyl", "target": "C_me", "order": 1},
    {"source": "C_acyl", "target": "C2", "order": 1},
    {"source": "N1", "target": "C2", "order": 1},
    {"source": "C2", "target": "C3", "order": 1},
    {"source": "C3", "target": "C4", "order": 1},
    {"source": "C4", "target": "C5", "order": 1},
    {"source": "C5", "target": "N1", "order": 1},
    {"source": "C3", "target": "O_me", "order": 1},
    {"source": "O_me", "target": "C_ome", "order": 1}
  ],
  "charges": null,
  "aromatic_n_h": null
}
\end{lstlisting}

\vspace{0.5em}
\centering
\includegraphics[width=0.50\linewidth]{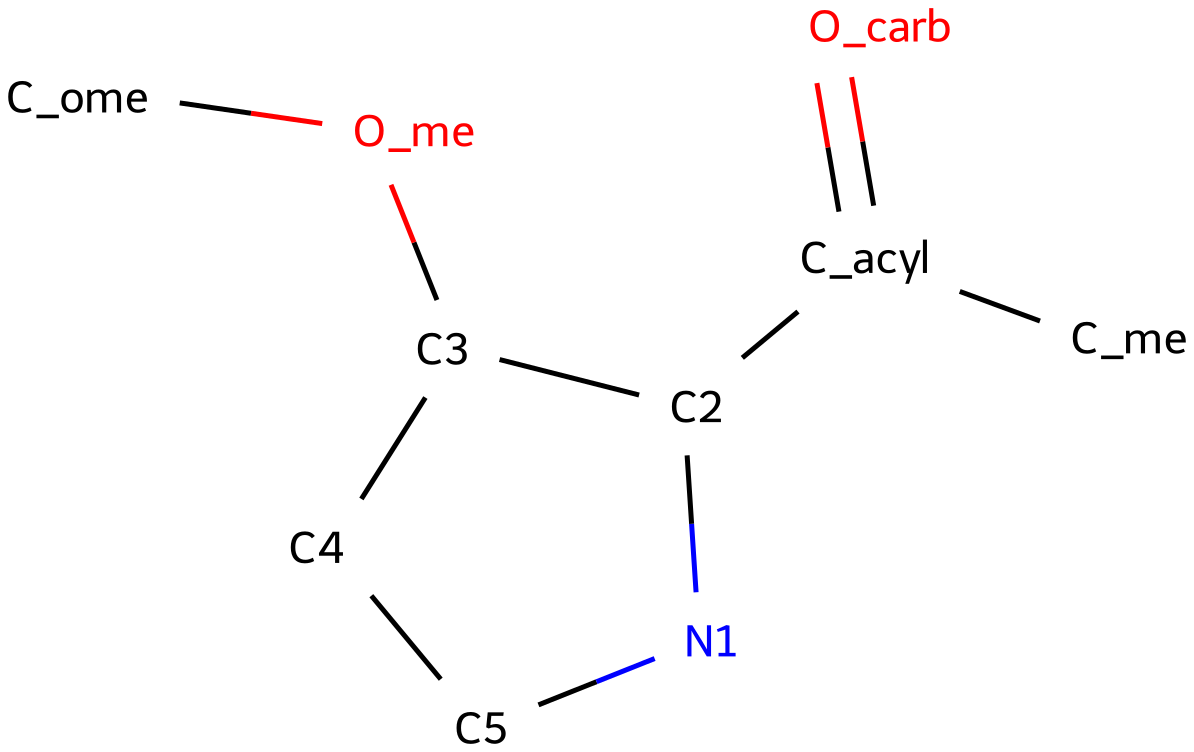}
  \end{tcolorbox}
  \caption{Example MolJSON generated by GPT-5 for an IUPAC to MolJSON translation question. The input IUPAC name was \emph{1-(3-methoxypyrrolidin-2-yl)ethanone}. The MolJSON which was generated by GPT-5 is shown in the figure, along with a visualisation of the corresponding molecule with the atoms showing the atom identifiers chosen by the model. In many cases, particularly for IUPAC to MolJSON questions, the model used rich atom identifiers, such as IDs that corresponded to functional group abbreviations.}
  \label{fig:graph-json-example}
\end{figure}

\begin{figure}[H]
  \centering
  \begin{tcolorbox}[
    colback=white,
    colframe=black,
    boxrule=0.5pt,
    arc=0pt,
    left=4pt,
    right=4pt,
    top=4pt,
    bottom=4pt
  ]
\begin{lstlisting}[
  basicstyle=\ttfamily\footnotesize,
  frame=none,
  breaklines=true,
  breakatwhitespace=false,
  showstringspaces=false
]
{
  "atoms": [
    {"id": "a1", "element": "N"}, 
    {"id": "a2", "element": "N"}, 
    {"id": "a3", "element": "C"}, 
    {"id": "a4", "element": "C"}, 
    {"id": "a5", "element": "C"}, 
    {"id": "a6", "element": "C"}, 
    {"id": "a7", "element": "C"}, 
    {"id": "a8", "element": "C"}, 
    {"id": "a9", "element": "C"}, 
    {"id": "a10", "element": "C"}, 
    {"id": "a11", "element": "O"}
  ], 
  "bonds": [
    {"source": "a1", "target": "a2", "order": 1.5}, 
    {"source": "a2", "target": "a3", "order": 1.5}, 
    {"source": "a3", "target": "a4", "order": 1.5}, 
    {"source": "a4", "target": "a5", "order": 1.5}, 
    {"source": "a5", "target": "a1", "order": 1.5}, 
    {"source": "a2", "target": "a7", "order": 1}, 
    {"source": "a7", "target": "a8", "order": 1}, 
    {"source": "a5", "target": "a6", "order": 1}, 
    {"source": "a4", "target": "a9", "order": 1}, 
    {"source": "a9", "target": "a10", "order": 1}, 
    {"source": "a10", "target": "a11", "order": 1}
  ], 
  "charges": [
    {"atom_id": "a2", "formal_charge": 1}
  ], 
  "aromatic_n_h": [
    {"atom_id": "a1", "hcount": 1}
  ]
}
\end{lstlisting}

\vspace{0.5em}
\centering
\includegraphics[width=0.7\linewidth]{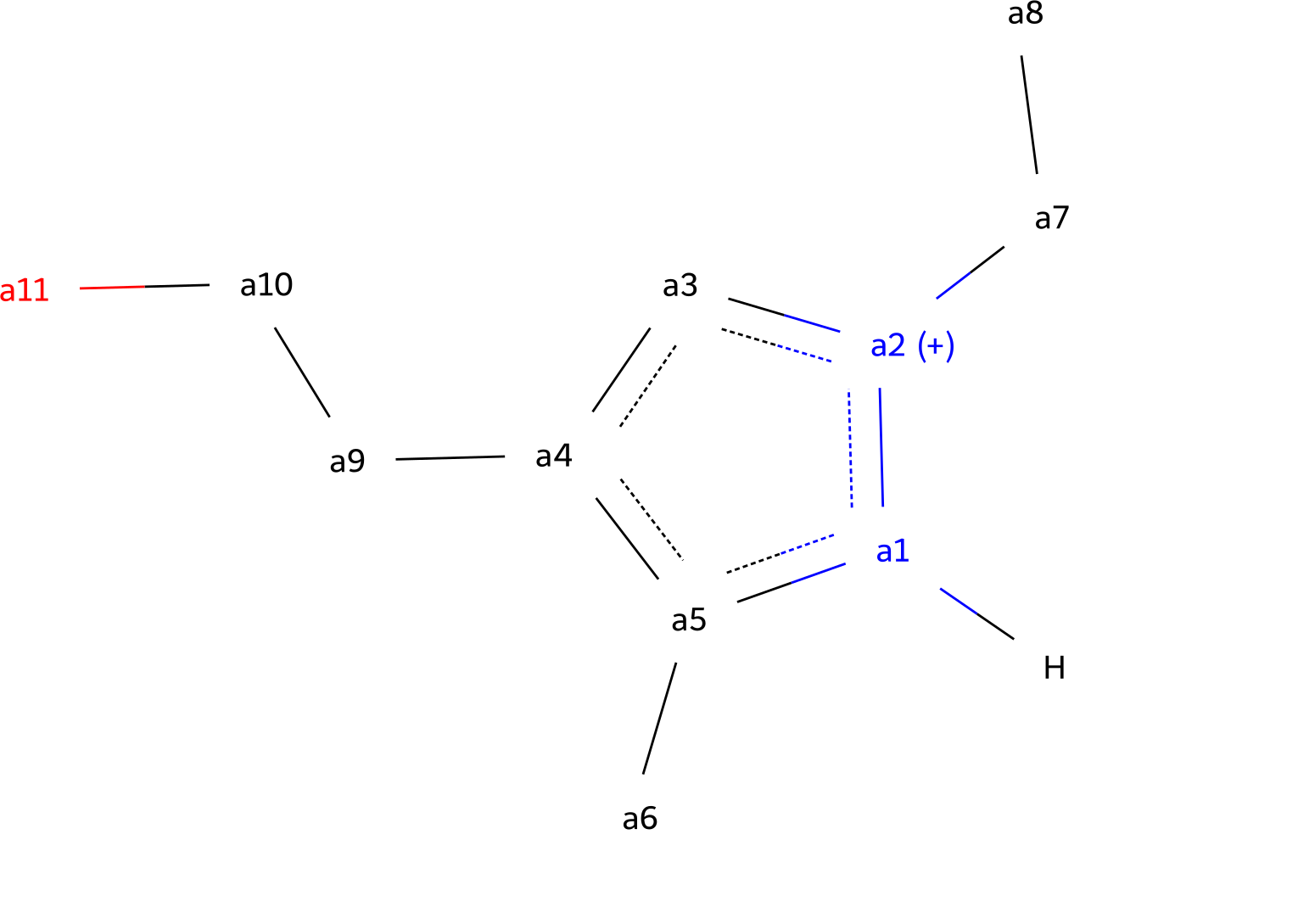}
  \end{tcolorbox}
  \caption{Example MolJSON generated by GPT-5 for an IUPAC to MolJSON translation task, where the molecule included a proton on an aromatic nitrogen, and an explicit charge. The input IUPAC name was \emph{2-(2-ethyl-5-methyl-1H-pyrazol-2-ium-4-yl)ethanol}. The MolJSON generated by GPT-5 is shown, along with a visualisation of the corresponding molecule with atom identifiers annotated.}
  \label{fig:MolJSON-charged-viz-example}
\end{figure}

\clearpage
\section{Analysis of Atom Identifiers Generated by GPT-5}
\label{sec:atom_id_analysis}

\begin{figure}[H]
  \centering
  \includegraphics[width=0.7\linewidth]{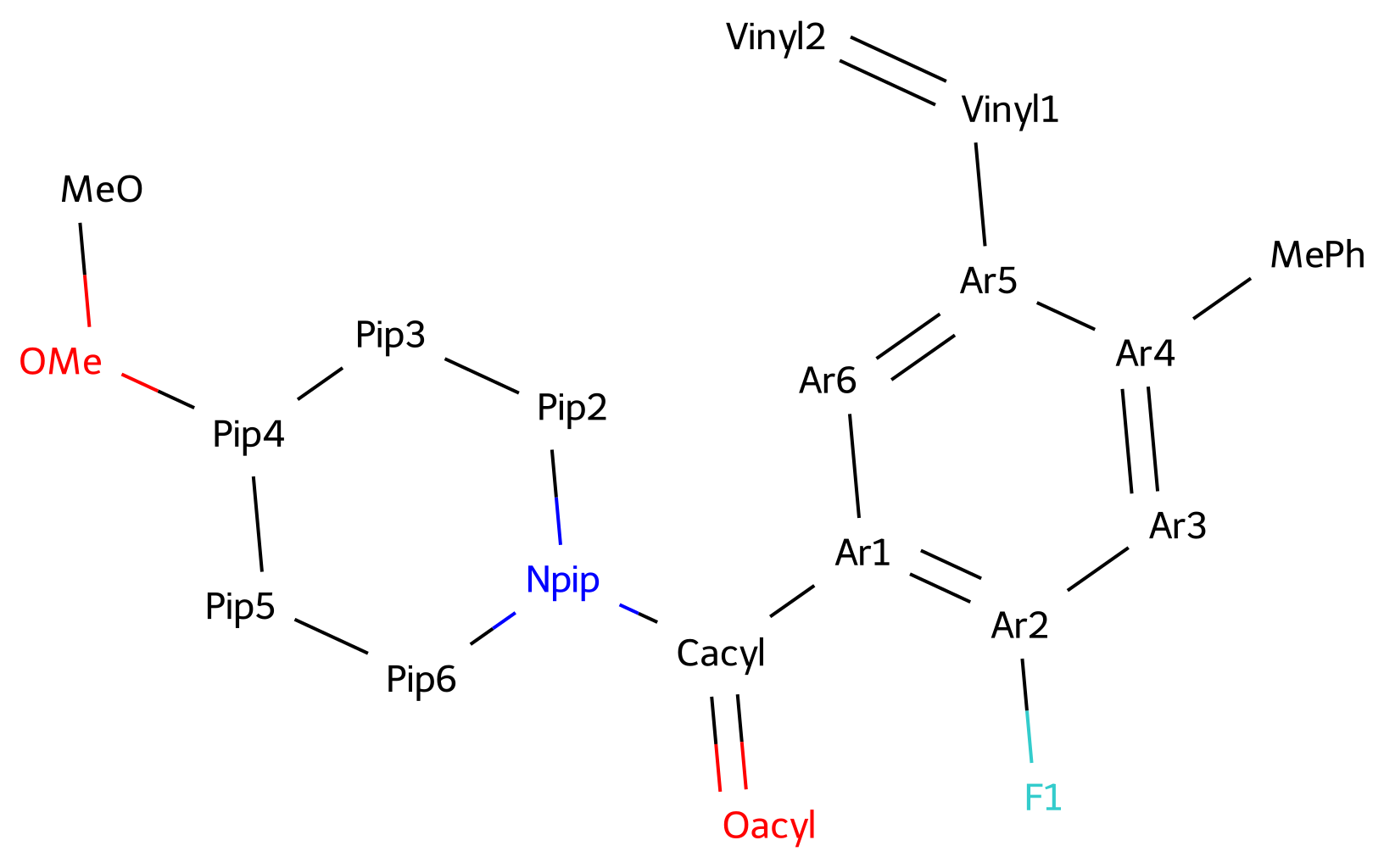}
  \caption{Example of a MolJSON output which uses rich atom identifiers. Each atom is annotated with the atom identifier that was selected by GPT-5. The choice of atom identifiers was spontaneous without explicit prompting; the only requirement in the MolJSON specification is that each atom ID is unique.}
  \label{fig:graph-json-rich-viz-example}
\end{figure}

\begin{figure}[H]
  \centering
  \includegraphics[width=\linewidth]{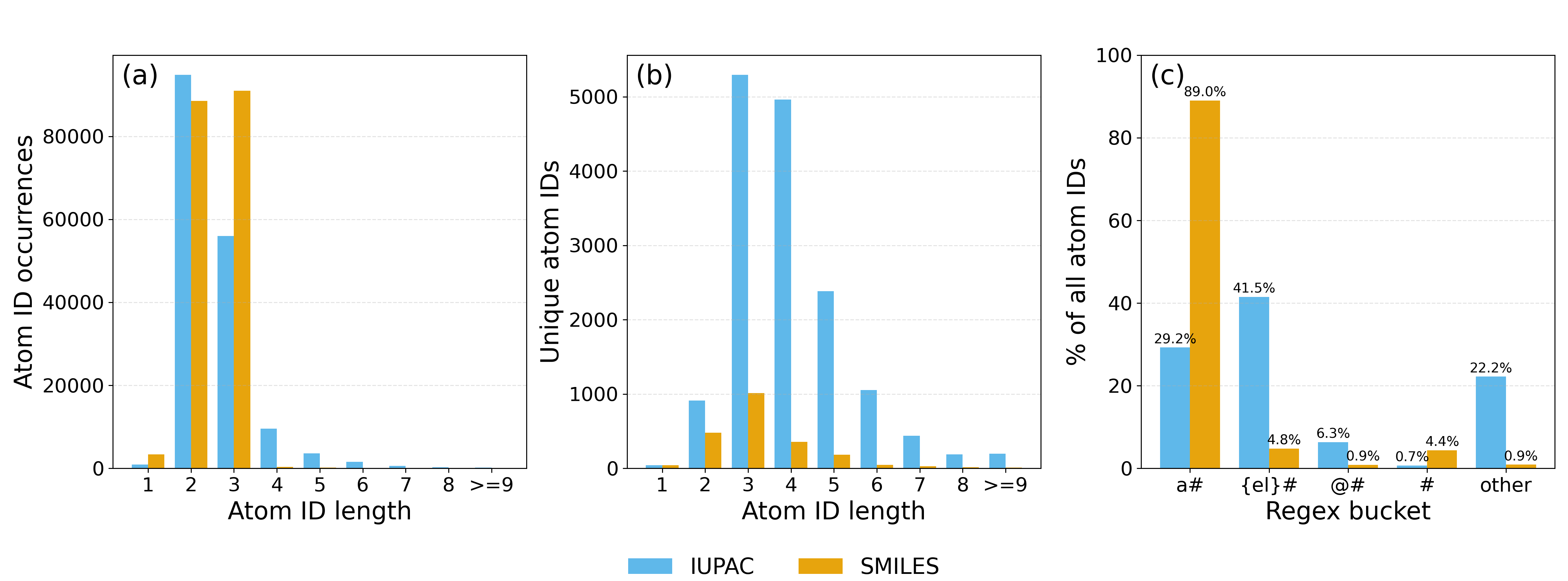}
  \caption{Distribution of atom identifiers used by GPT-5 in MolJSON outputs. (a) Total atom ID occurrences by string length. (b) Unique atom ID counts by string length. (c) Crude regex binning of atom identifiers into five buckets: ``a\#'' -- the letter 'a' followed by an integer; ``\{el\}\#'' -- any element symbol followed by an integer; ``@\#'' -- any singular letter followed by an integer; ``\#'' -- any integer; ``other'' -- any uncategorised atom IDs, a sample of which are visualised in Appendix \ref{sec:atom_id_analysis} Figure \ref{fig:atomid-word-cloud}. The distribution of atom IDs generated by GPT-5 varied depending on the input representation, with IUPAC inputs resulting in a more diverse and rich selection of atom IDs.}
  \label{fig:atomid-length-and-waterfall}
\end{figure}

We propose that part of the reason for the high output accuracy of MolJSON is due to the format being more expressive. 
In particular, the LLM is allowed to choose any unique identifier to refer to each atom in the molecule. 
When exploring the atom IDs selected by the LLMs for SMILES to MolJSON and IUPAC to MolJSON, we observed the models chose markedly different identifiers in each case (Appendix \ref{sec:atom_id_analysis} Figure \ref{fig:atomid-length-and-waterfall}). 
For SMILES to MolJSON translation, the majority of atom IDs followed a simple numbering structure of the form \texttt{a1, a2, a3...}. 
In contrast, the IUPAC to MolJSON atom IDs most often followed a numbering system using elements \texttt{C1, C2, O1...}, and in around 22.1\% cases the atom ID was a rich string, which sometimes corresponded to functional group abbreviations (Appendix \ref{sec:atom_id_analysis} Figure \ref{fig:graph-json-rich-viz-example}).
We hypothesise that rich identifiers could be functionally used as ``graph tokens'' by the LLM within the Chain-of-Thought, and that the permissive MolJSON output format can allow the LLM to emit these tokens directly. 
This is in contrast to IUPAC naming, where the graph needs to be converted to language using nomenclature rules, and SMILES, where a traversal is required to convert the graph into SMILES. 
It is possible that training an LLM to reason using an explicit graph representation, such as MolJSON or otherwise, could improve performance across other chemical reasoning tasks.

\begin{figure}[H]
  \centering
  \includegraphics[width=\linewidth]{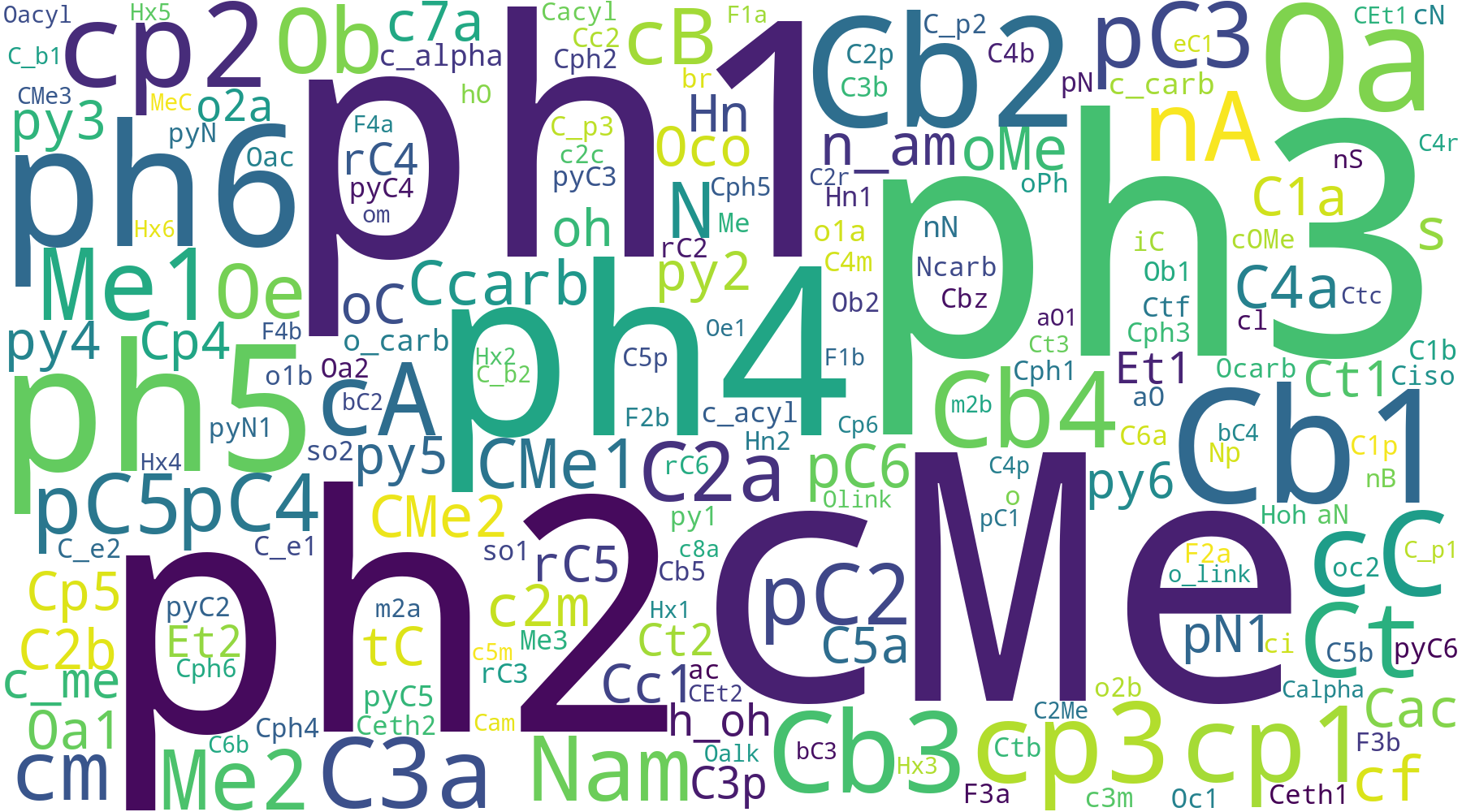}
  \caption{Word cloud of the top 300 most frequent atom identifiers generated by GPT-5 for IUPAC to MolJSON questions that were classified as ``other'' in Appendix \ref{sec:atom_id_analysis} Figure \ref{fig:atomid-length-and-waterfall}.}
  \label{fig:atomid-word-cloud}
\end{figure}

\end{document}